\documentclass[letterpaper]{article} 
\usepackage{aaai23}  
\usepackage{times}  
\usepackage{helvet}  
\usepackage{courier}  
\usepackage[hyphens]{url}  
\usepackage{graphicx} 
\usepackage{natbib}  
\usepackage{caption} 
\usepackage{microtype}
\usepackage{booktabs} 

\usepackage{hyperref}

\usepackage{amsmath}
\usepackage{amssymb}
\usepackage{mathtools}
\usepackage{amsthm}
\usepackage{graphicx} 

\usepackage{algorithm}
\usepackage{algorithmic}
\usepackage{pgfplots}
\usepackage[labelformat=simple]{subcaption}
\usepackage{mdframed}
\usepackage{adjustbox}
\usepackage{multirow}
\usepackage[capitalize,noabbrev]{cleveref}
\usepackage{dblfloatfix}
\usepgfplotslibrary{groupplots}
\usepackage{xspace}
\usetikzlibrary{backgrounds} 

\usetikzlibrary{calc,shapes.geometric,shapes,automata,fit}
\usepgfplotslibrary{fillbetween}

\usepackage{newfloat}
\usepackage{listings}
\floatstyle{ruled}
\newfloat{listing}{tb}{lst}{}
\floatname{listing}{Listing}

\newtheorem{definition}{Definition}
\newtheorem{problem}{Problem}

\newtheorem{theorem}{Theorem}

\newcommand{\ie}{i.\,e.,\@\xspace}

\newcommand{\eg}{e.\,g.,\@\xspace}

\newcommand{\Act}{\ensuremath{\mathit{Act}}}
\newcommand{\act}{\ensuremath{a}}

\newcommand{\Distr}{\mathit{Distr}}

\DeclareMathOperator{\supp}{supp}

\newcommand{\ObsSym}{{Z}}
\newcommand{\ObsFun}{{O}}
\newcommand{\obs}{\ensuremath{z}}
\newcommand{\pomdp}{\mathcal{M}}
\newcommand{\states}{\ensuremath{S}}
\newcommand{\init}{\ensuremath{I}}
\newcommand{\probmdp}{\mathcal{P}}
\newcommand{\rew}{\ensuremath{\mathcal{R}}}
\newcommand{\osched}{\ensuremath{\mathit{\pi}}}

\newcommand{\belief}{\mathfrak{b}}
\newcommand{\belsup}{B}
\newcommand{\belsups}{\mathcal{B}}
\newcommand{\avoid}{\textsl{AVOID}}
\newcommand{\reach}{\textsl{REACH}}

\newcommand{\psched}{\nu}
\newcommand{\estimator}{\sigma}

\newcommand{\probability}{\mathit{Pr}}

\definecolor{steve_red}{HTML}{FF0F19}
\definecolor{randomgray}{RGB}{128,128,128}
\definecolor{steve_green}{RGB}{1,68,33}
\definecolor{steve_yellow}{HTML}{FDE9CA}
\definecolor{steve_highlight}{HTML}{E99076}

\newcommand{\tubeplotindex}[4]{\addplot[name path=A,draw=none] table[x index={0}, y index={#3},forget plot] from #2;
\addplot[name path=B,draw=none] table[x index={0}, y index={#4},forget plot] from #2;
\addplot[#1,opacity=0.1,forget plot] fill between[of=A and B];
\addplot[#1,line width=0.04cm] table[x index={0}, y index={1}] from #2;}

\newcommand{\domain}[1]{\textsl{#1}}

\newcommand{\screenshotsize}{1.25in}

\newcommand{\rlgraphheight}{0.21*\textwidth}

\usepackage[textsize=tiny]{todonotes}
\title{Safe Reinforcement Learning via Shielding under Partial Observability
}

\author {
	Steven Carr,\textsuperscript{\rm 1}
	Sebastian Junges, \textsuperscript{\rm 2}
	Nils Jansen\textsuperscript{\rm 2}
	Ufuk Topcu,\textsuperscript{\rm 1}
}
\affiliations {
	\textsuperscript{\rm 1} The University of Texas at Austin
	\textsuperscript{\rm 2} Radboud University, Nijmegen, The Netherlands\\
}
\begin{document}
\maketitle

\begin{abstract}
Safe exploration is a common problem in reinforcement learning (RL) that aims to prevent agents from making disastrous decisions while exploring their environment. 
A family of approaches to this problem assume domain knowledge in the form of a (partial) model of this environment to decide upon the safety of an action. 
A so-called shield forces the RL agent to select only safe actions. 
However, for adoption in various applications, one must look beyond enforcing safety and also ensure the applicability of RL with good performance. 
We extend the applicability of shields via tight integration with state-of-the-art deep RL, and provide an extensive, empirical study in challenging, sparse-reward environments under partial observability. 
We show that a carefully integrated shield ensures safety and can improve the convergence rate and final performance of RL agents. 
We furthermore show that a shield can be used to bootstrap state-of-the-art RL agents: they remain safe after initial learning in a shielded setting, allowing us to disable a potentially too conservative shield eventually.

\end{abstract}
\section{Introduction}

\noindent Reinforcement learning (RL)~\citep{sutton1998reinforcement} is a technique for decision-making in uncertain environments.
An \emph{RL agent} explores its environment by taking \emph{actions} and perceiving feedback signals, usually \emph{rewards} and \emph{observations} on the system state.
With success stories such as
AlphaGo~\citep{DBLP:journals/nature/SilverHMGSDSAPL16} 
RL nowadays reaches into areas such as robotics~\citep{DBLP:journals/ijrr/KoberBP13} or autonomous driving~\citep{DBLP:journals/corr/SallabAPY17}.

A significant limitation of RL in safety-critical environments is the high cost of failure.
An RL agent explores the effects of actions  -- often selected randomly, such as in state-of-the-art policy-gradient methods~\citep{DBLP:conf/iros/PetersS06} -- and will thus inevitably select actions that potentially cause harm to the agent or its environment. 
Thus, typical applications for RL are games~\citep{DBLP:journals/corr/MnihKSGAWR13} or assume the ability to learn on high-fidelity simulations of realistic scenarios~\citep{DBLP:journals/tii/TaoZLN19}. 
The problem of \emph{unsafe exploration} has triggered research on the \emph{safety} of RL~\citep{garcia2015comprehensive}.
Safe RL may refer to (1)~changing (``\emph{engineering}'') the reward function~\citep{laud2004theory} to encourage the agent to choose safe actions, (2)~adding a second cost function (``\emph{constraining}'')~\citep{DBLP:conf/icml/MoldovanA12}, or (3)~blocking (``\emph{shielding}'') unsafe actions at runtime~\citep{shield_rl}. 
This paper falls into the last category.

Safe RL in partially observable environments suffers from uncertainty both in the agent's actions and perception.
Such problems, typically modeled as partially observable Markov decision processes (POMDPs)~\citep{kaelbling1998planning}, require histories of observations to extract a sufficient understanding of the environment.
Recent deep RL approaches for POMDPs, including those that employ recurrent neural networks ~\citep{hausknecht2015deep,wierstra2007solving}, learn from these histories and can generate high-quality policies with sufficient data.
However, these approaches do not guarantee safety during or after learning.  
While shielding for Markov decision processes (MDPs) is rather well-studied~\cite{shield,shield_rl,DBLP:conf/aaai/FultonP18,DBLP:journals/corr/abs-1904-07189}, there is---to the best of our knowledge---no approach that integrates shielding with deep RL. 


%

We contribute a thorough study, implementation, and experimental evaluation on integrating state-of-the-art RL for POMDPs with shields.
We demonstrate effects and insights using several typical examples and provide detailed information on RL performance as well as videos showing the exploration and training process.
In particular, we integrate various RL algorithms from Tensorflow~\citep{TFAgents} with a shield that guarantees safety regarding so-called reach-avoid specifications, a special case of temporal logic constraints~\cite{pnueli1977temporal}.
The computation of the shields is based on satisfiability solving~\citep{DBLP:conf/cav/JungesJS20} and requires only a \textit{partial model of the environment}.
Specifically, we need to know all potential transitions in the POMDP, while probabilities and rewards may remain unspecified~\citep{DBLP:journals/lmcs/RaskinCDH07}.

\begin{figure}[t]
    \centering
    \scalebox{0.92}{
    \begin{tikzpicture}
       \node[draw, minimum width=2cm,fill=black!10] (agent) { Agent};
       \node[right=2cm of agent, draw, minimum width=2cm,fill=black!10] (env) { Environment};
       
       \node[very thick, draw, below=2.5cm of agent,xshift=-1em] (filter) {State estimator};
       \node[very thick, draw, below=0.4cm of filter] (shield) {Shield};
       \begin{scope}[on background layer]
       \node[draw, fit=(agent)(env), minimum height=2.5cm, inner sep=7pt,fill=black!4] (rlenv) {};
       \end{scope}
       
       \draw[->] (agent.north) |- ($(agent.north)+(0,0.4)$)  -| node[above,pos=0.25] {action} (env.north);
       \draw[->] (env.south) |- ($(env.south)+(0,-0.6)$)  -| node[pos=0.25,above=-0.0cm,text width=1.75cm] {observation \& reward} (agent.south);

       \node[ellipse,right=2.0cm of filter.350,draw,xshift=-0.5em] (graph) { Partial model};
       \node[right=0.0cm of env,ellipse,draw,yshift=-4.5em] (model) { Model};
       
       \draw[->,dotted, thick] (env) -| node[above=-0.5cm,xshift=-0.15em,text width=1.0cm,align = right] { described by} (model);
       \draw[->,dotted, thick] (model) -| node[right=0.0cm,pos=0.6,text width=2.5cm,yshift=-1.35em] {abstract} (graph.north);

         \node[ellipse,below=0.15cm of graph,draw] (spec) { Safety spec};
       
         \draw[dashed,->] (agent.west) |- ($(agent.west)+(-1.05,0)$)  |- node[pos=0.25,left] {\rotatebox{90}{queries}} (filter.west);
       \draw[dashed,->] (agent.west) |- ($(agent.west)+(-1.05,0)$)  |-  (shield.west);
       \draw[thick, dotted,->] (spec) -- node[below,pos=0.27] {} (shield.345);

       \draw[thick, dotted,->] (graph) -- node[pos=0.5,below]{} (filter.350);
       \draw[thick, dotted,->] ($(graph.west)+(-1.25,0)$) |- (shield.0);
       
       \draw[->] (rlenv.270) |- node [right=0.05cm,pos=0.1,text width=1.75cm,yshift=-0.75em] { observation \& action}(filter.10);
       \draw[->] (rlenv.270) |- (shield.15);
       \begin{scope}[on background layer]
       \node[draw, fit=(filter)(shield), minimum height=2.0cm, inner sep=7pt,fill=black!4,label=above:{Knowledge interface}] (rlenv) {};
       \end{scope}
       
    \end{tikzpicture}}
    \caption{Safe RL with two knowledge interfaces for the agent: A state estimator and a shield, based on a partial model of the environment.}
    \label{fig:outline}
\end{figure}
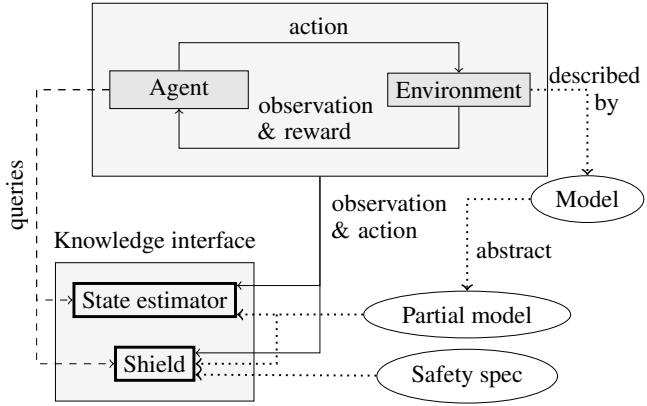

\paragraph{Approach.}
Fig.~\ref{fig:outline} shows the outline of the safe RL setting.
The gray box shows a typical RL procedure.
The environment in the partially observable setting is described by a POMDP model that may not be fully known.
A \textit{shield} in this setting (implicitly) requires access to a form of \textit{state estimation} to account for a \textit{safety specification}.
This estimator uses the \textit{partial model} of the environment to track in which states the model may be, based on the observed history.
We see the shield and the state estimator as two knowledge interfaces for the agent that may be used in conjunction or separately.
A shield may be too conservative and overly protective, and therefore 
restrict the performance of RL in general. 
Alternatively, we investigate if (only) access to a state estimator may serve as a lightweight alternative to a shield.
We show that, while the RL agent may indeed benefit from this additional information, the shield provides more safety and faster convergence than relying on just the state estimator.
After learning, we may gently phase out a shield and still preserve the better performance of the shielded RL agent. 
Then, even an overly protective shield helps to bootstrap RL.

\newcommand*{\connectorH}[4][]{
  \draw[#1] (#3) -| ($(#3) !#2! (#4)$) |- (#4);
}
\newcommand*{\connectorV}[4][]{
  \draw[#1] (#3) |- ($(#3)+(0,#2)$) -| (#4);
}

Summarized, our study demonstrates the following effects of shielding in a partially observable setting. 
\begin{itemize}
	\item \emph{Safety during learning:} Exploration is only safe when the RL agent is provided with a shield. 
	Without the shield, the agent makes unsafe choices even if it has access to the state estimation. 
	Even an unshielded \emph{trained agent} still behaves unsafe sometimes.
	\item \emph{RL convergence rate:} A shield not only ensures safety, but may also significantly improve the convergence rate of modern RL agents by avoiding spending time to learn unsafe actions. 
	Other knowledge interfaces like state estimators do help to a lesser extent.
	\item \emph{Bootstrapping:} Due to the improved convergence rate, shields are a way to bootstrap RL algorithms, even if they are overly restrictive. 
	RL agents can learn to mimic the shield by slowly disabling the shield.
\end{itemize}
\paragraph{Further related work.}
Several approaches to safe RL in combination with formal verification exist~\citep{DBLP:conf/atal/HasanbeigAK20,shield,shield_rl,jansen-et-al-concur-2020,DBLP:conf/aaai/FultonP18,DBLP:journals/corr/abs-1904-07189}.
These approaches either rely on shielding, or guide the RL agent to satisfy temporal logic constraints.
However, none of these approaches take our key problem of partial observability into account.
Recent approaches to find safe policies for POMDPs with partial model knowledge either do not consider reinforcement learning~\citep{DBLP:conf/aaai/Cubuktepe0JMST21} or require the agent to take catastrophic actions before learning from them~\cite{DBLP:journals/corr/abs-2202-09516}.


\section{Problem Statement}

We introduce POMDPs as the standard model for sequential decision-making under partial observability.
We distinguish the learning goal of an agent by expected rewards, and the safety constraints via reach-avoid safety specifications.

\subsection{POMDPs}
A (discrete) \emph{partially observable Markov decision process (POMDP)} is a tuple $\pomdp = (\states,\init,\Act, \ObsFun,\ObsSym, \probmdp, \rew)$ where $\states$ is a finite state space.
$I$ is the initial distribution that gives 
the probability~$I(s)$ that the agent starts in state $s\in\states$,
 and $\Act$ is a finite space of actions.
 $\ObsSym$ is a finite observation space and $\ObsFun(\obs|s)$ is the probability of observing $\obs$ when the environment is in state $s$.
The transition model $\probmdp(s'| s,\act)$ represents the conditional probability of moving to a state $s'\in S$ after executing $a\in A$ in $s\in S$.
When executing $\act\in\Act$ in $s\in \states$, the agent receives a scalar reward $\rew(s,\act)$.
As not every action may be available in every state,
we define the set of available actions in $s$ as $\Act(s)$.
We remark that POMDPs may have dead-ends from which an agent cannot obtain positive rewards~\citep{DBLP:conf/uai/KolobovMW12}.

As the {current} state of a POMDP is not observable, agents may infer the probability of being in a certain state based on the history so far.
This probability distribution is the \textit{belief} $\belief\colon (\ObsSym\times \Act)^*\times \ObsSym \rightarrow Distr(S)$.
A \emph{policy} $\pi\colon \belief\rightarrow\Distr(\Act)$ for the agent decides upon a distribution over actions based on the current belief.
A (fully observable) \emph{belief MDP} with the (infinitely many) beliefs of the POMDP as states captures the belief dynamics.
%

\subsection{Learning Goal and Safety Constraints}
The standard \emph{learning goal} for POMDPs is to find a policy $\osched$ that maximizes the expected discounted reward, that is, $\mathbb{E}\left[\sum_{t=0}^\infty\gamma^t\rew_t\right]$, where $\gamma^t$ with $0\leq \gamma^t\leq 1$ is the discount factor and $\rew_t$ is the reward at time $t$.	
%
Due to the infinite number of beliefs, computing such an optimal policy 
is in general undecidable, even if the entire model is known~\citep{MadaniHC99}.

%


An agent in safety-critical settings must (additionally) adhere to \textit{safety constraints}. We capture these constraints using  \emph{(qualitative) reach-avoid} specifications, a subclass of indefinite horizon properties~\citep{Put94}.
Such specifications necessitate to \emph{always} avoid certain bad states from $\avoid\subseteq \states$ and reach states from $\reach\subseteq\states$ \emph{almost-surely}, i.e., with probability one (for arbitrary long horizons).
We denote these constraints by  $\varphi=\langle\reach,\avoid\rangle$.
The \emph{avoid} specification $\varphi=\langle\avoid\rangle$ necessitates only to avoid bad states.
Formally, $\probability_\belief^\osched(S)$ denotes the probability to reach a set of states $S'\subseteq S$ from the belief $\belief$ using the policy $\osched$.

\begin{definition}[Winning]
	A policy $\osched$ is \emph{winning} for specification $\varphi$ from belief $\belief$ in POMDP $\pomdp$ iff $\probability_\belief^\osched(\avoid)=0$ and $\probability_\belief^\osched(\reach)=1$.
	%
	Belief $\belief$ is \emph{winning} for $\varphi$ in $\pomdp$ if there exists a winning policy from~$\belief$.
\end{definition}
The relation $\pomdp(\osched)\models\varphi$ denotes that the policy $\osched$ is winning
according to the initial belief $I$.
Computing such a \emph{winning} policy is, in general, decidable and EXPTIME-complete but practically feasible methods exist~\citep{DBLP:conf/aaai/ChatterjeeCD16,DBLP:conf/cav/JungesJS20}.
Finally, for multiple beliefs, a \emph{winning regions} (aka safe or controllable regions) is a set of winning beliefs, that is, from each belief within a winning region, there exists a winning policy.



We formulate the joint learning and safety problem we consider in our safe reinforcement learning setting. 
\begin{mdframed}
\begin{problem}[Safe Learning]\label{problem:max-rew-reach-avoid}
Given a POMDP $\pomdp$, a safety constraint $\varphi$, and let $\osched_1,\ldots,\osched_n$ be the  (training) sequence of policies employed by an RL agent. 
The problem is to ensure that for all policies $\osched_i$ it holds that $\pomdp(\osched_i)\models\varphi$ with $1\leq i\leq n$ and the final policy $\pi_n$ maximizes $\mathbb{E}\left[\sum_{t=0}^\infty\gamma^t\rew_t\right]$.
\end{problem}
\end{mdframed}
Note that the condition to satisfy $\varphi$ may induce a sub-optimal reward as the agent has to strictly adhere to the safety constraint while collecting rewards.

\section{State Estimators and Shields}
In this section, we present the two knowledge interfaces for RL agents in the shielding approach for POMDPs, refer to Fig.~\ref{fig:outline}.
In particular, we discuss belief supports as concrete state estimators for the agent, and introduce the notion of a shield (for POMDPs).
We discuss which guarantees we can provide for shields that are computed on partial models.

\subsection{Belief Supports as State Estimators}
If the transition (and observation) probabilities in the POMDP were known, the agent could incrementally compute a Bayesian update of the belief and use that to estimate its current state.
However, this is a strong assumption.
Instead, we rely on the so-called belief support. 
A state $s$ is in the \emph{belief support} $\belsup$
for a belief $\belief$, if  $s\in\supp(\belief)$.
Thus, the belief-support $\belsup \in \belsups$, with $\belsups$ the set of all belief supports, is a set of states.
The belief-support can be updated using only the graph of the POMDP (without probability knowledge) by a simplified belief update. In particular, we can compute the unique belief support $(\belsup' \mid  \belsup, \act, \obs)$ that can be reached from $\belsup$ with action $\act$ and observation $\obs$. 
We exploit the following result from~\cite{DBLP:conf/cav/JungesJS20}.
\begin{theorem}[]\label{theo:belsup}
If a belief $\belief$ is winning for a reach avoid specification $\varphi$, any belief $\belief'$ with $\supp(\belief')=\supp(\belief)$ is winning.
\end{theorem}
\noindent Intuitively, all beliefs that share a belief-support are winning, therefore, we can directly call a belief-support winning.
We now define the first knowledge interface for the RL agent.

\paragraph{Knowledge interface 1: the state estimator.}
A \textit{belief-support state estimator} $\estimator \colon (\ObsSym\times \Act)^*\times \ObsSym \rightarrow \belsups$ takes as input the observation-action history and returns the current belief support to the RL agent. 
This estimator can be implemented by repeatedly updating the belief-support independently of the probabilities in a POMDP.
We provide the agent with state estimation as additional observation signal.

\subsection{Shields}

The purpose of a shield is to prevent the agent from taking actions that would violate a (reach-avoid) specification.
A shield allows only actions that enable the agent to stay in a winning region.
That means, when taking an action from some winning  belief support, any next belief support reached by taking this action must belong to a winning region.




We first define policies on the belief support of the form $\osched\colon \belsups\rightarrow \Act$.
Recall that $\belsups$ denotes the set of all belief supports.
This (\emph{deterministic}) policy chooses one unique action for each belief support $\belsup \in \belsups$.
For shields, we use a more liberal notion of \emph{permissive} policies that select sets of actions~\citep{DBLP:journals/corr/DragerFK0U15,DBLP:conf/tacas/Junges0DTK16}.
Given a POMDP $\pomdp$, a permissive policy is given as $\psched\colon \belsups \rightarrow 2^\Act$.
Any action in $\psched(\belsup)$ is called \emph{allowed}.
Intuitively, one can think of a permissive policy as defining a set of policies: 
In particular, a  policy $\osched$ is \emph{admissible} for $\psched$ if for all belief supports $\belsup \in \belsups$ it holds that $\osched(\belsup) \in \psched(\belsup)$\footnote{Admissibility can be defined for more general classes of policies, see \cite{DBLP:conf/cav/JungesJS20}.}.

Shields can be defined as permissive policies for staying in a winning region with respect to reach-avoid specifications. 
\begin{definition}[Shield]
For a specification $\varphi$, a permissive policy $\psched$ is a $\varphi$-shield, if for any $\belsup$ winning for $\varphi$,  
 $\act \in \psched(\belsup)$,
 and $\obs \in \ObsSym$, 
 it holds that $(\belsup' \mid  \belsup, \act, \obs)$ implies $\belsup'$ is winning.
\end{definition}

Shields provide guarantees such that any policy that agrees with the shield is winning~\cite{DBLP:conf/cav/JungesJS20}. 
We first state the formal correctness for avoid specifications.
\begin{theorem}
Let $\varphi$ be an avoid-specification.  
For any \emph{$\varphi$-shield for $\pomdp$} all  admissible policies are winning.
\end{theorem}

Shields for avoid-specifications may block all actions and create deadlocks. 
Instead, we employ shields
for reach-avoid specifications that prevent the agent from  visiting any dead-ends.
A shield itself cannot force an agent to visit reach states. 
However, under mild assumptions, we can ensure that the agent eventually visits the reach states: A policy is \emph{fair} if in any state which is visited infinitely often, it also takes every allowed action infinitely often~\cite{BK08}.
For example, a policy that takes every allowed action with positive probability is fair.
\begin{theorem}
Let $\varphi$ be a reach-avoid-specification.  
For a \emph{$\varphi$-shield for $\pomdp$}, all fair and admissible policies are winning.
\end{theorem}

\paragraph{Knowledge interface 2: the shield.}
 We use shields as computed via~\cite{DBLP:conf/cav/JungesJS20} that ensure reach-avoid specifications as outlined above.
 Essentially, an agent may use a state estimator $\estimator \colon (\ObsSym\times \Act)^*\times \ObsSym \rightarrow \belsups$ in conjunction with a shield $\psched\colon \belsups \rightarrow 2^\Act$ to compute allowed actions.
 We restrict the available actions for the RL agents to these allowed actions.




\subsection{Shields on Partial Models}
A shield for a POMDP relies only on the belief-support.
Therefore, it is also a shield for all POMDPs with the same underlying graph-structure.
We formalize this statement.

\paragraph{Partial models.}
We assume the agent has only access to a \emph{graph-preserving approximation} $\pomdp'$ of a POMDP $\pomdp$ that differs only in the transition models $\probmdp'$ and $\probmdp$, and potentially in the reward functions.
It holds that $\probmdp(s'| s,\act)>0 \iff \probmdp'(s'| s,\act)>0$ for all states $s,s'\in \states$ and actions $\act\in\Act$.
Similarly, $\pomdp'$ \emph{overapproximates the transition model of $\pomdp$}, if it holds for all states $s,s'\in \states$ and actions $\act\in\Act$ that $\probmdp(s'| s,\act)>0 \implies \probmdp'(s'| s,\act)>0$.
The original POMDP has no transitions that are not present in the partial model.

\paragraph{Guarantees.} We state the following guarantees provided by a shield, in relation to the two approximation types.
\begin{theorem}[Reach-Avoid Shield]\label{theo:ra-shield}
 Let $\pomdp'$ be a graph-preserving approximation of $\pomdp$ and $\varphi$
a reach-avoid specification, then a $\varphi$-shield for $\pomdp'$ is a $\varphi$-shield for $\pomdp$.
\end{theorem}
This theorem follows directly from Theorem~\ref{theo:belsup}. 
Knowing the exact set of transitions with (arbitrary) positive probability for a POMDP is sufficient to compute a $\varphi$-shield. 


For avoid specifications, we can further relax the assumptions. It suffices to require that each transition in the partial model exists (with positive probability) in the (true) POMDP.
\begin{theorem}[Avoid Shield]\label{theo:a-shield}
Let $\pomdp'$ overapproximate the transition model of $\pomdp$, and let $\varphi'=\langle \avoid \rangle$ be an avoid specification, then a $\varphi'$-shield for the partial model $\pomdp'$ is a $\varphi'$-shield for the POMDP $\pomdp$.
\end{theorem}
If the partial model is further relaxed, it is generally impossible to construct a shield with the same hard guarantees. 

\section{RL with Partial Model Knowledge}
\label{sec:benefits}

Recall the general setup in Fig.~\ref{fig:outline}:
While the environment is described as a POMDP, the agent has only access to a partial model via the knowledge interfaces as explained in the previous section.
This section discusses the potential benefits of using these interfaces.

\subsection{Safety}
\paragraph{Safety during learning.}
Shielded RL agents are guaranteed to be safe during learning provided that the partial model is adequate as formalized by Theorem~\ref{theo:ra-shield}. 
Furthermore, assuming the RL agent is fair as defined above, it is guaranteed that they will eventually reach the $\reach$ states. This guarantee does not come with an upper bound on the number of steps\footnote{Shields can also be computed for finite-horizon or cost-bounded reach-avoid properties, which come with a guarantee on finite steps.}. 
In contrast, an unshielded agent takes actions first to learn that it may lead to an $\avoid$ state. 
State estimators are thus not sufficient to ensure safety, as they only reason about history and not about the future.


\paragraph{Safety after learning.}
In general, safety objectives encoded as reward and performance objectives (also encoded as reward) may allow for non-trivial trade-offs, which harm the ability to learn to adhere to safety objectives. Shielded RL agents do not face this tradeoff as they  must adhere to the explicit safety constraints.  


\paragraph{State estimators and safety.}
While state estimators cannot guarantee safety, they may improve safety. 
In particular, consider an action (such as switching on a light) which is useful and safe in most but not all situations (\eg a gas leak).
A state estimator provides the additional observation signals that allow the RL agent to efficiently distinguish these states, thereby indirectly improving safety, even during learning.


\begin{figure}
\centering
\renewcommand\thesubfigure{(\alph{subfigure})}
\subcaptionbox{Example for estimators 
    \label{fig:estimatorconvergence}}{
    \centering
    \scalebox{1.0}{
    \begin{tikzpicture}
     \node[circle,draw, initial, initial text=,initial where=above] (s1) {};
     \node[circle,draw,right=0.2cm of s1,yshift=1.4em, fill=orange] (s2) {};
     \node[circle,draw,right=0.2cm of s1,yshift=-1.4em, fill=blue] (s3) {};
     \node [cloud, draw,cloud puffs=10,cloud puff arc=120, aspect=2, inner ysep=0.5em,right=0.3cm of s2, fill=black!10] (c2) {};
     \node [cloud, draw,cloud puffs=10,cloud puff arc=120, aspect=2, inner ysep=0.5em,right=0.3cm of s3, fill=black!10] (c3) {};
     
     \node[circle,draw,right=0.4cm of c2,fill=black!10] (s4) {};
     \node[circle,draw,right=0.4cm of c3,fill=black!10] (s5) {};
     
     \node[draw, circle, right=0.5cm of s4,accepting] (acc) {};
     \node[draw, circle, right=0.5cm of s5] (rej) {};
     
     \draw[->] (s1) --node[above,xshift=-0.1cm] {\scriptsize $\textrm{up}$} (s2);
     \draw[->] (s1) --node[below,xshift=-0.2cm] {\scriptsize $\textrm{down}$} (s3);
     
     \draw[->] (s2) -- (c2);
     \draw[->] (s3) -- (c3);
     
     \draw[->] (c2) -- (s4);
     \draw[->] (c3) -- (s5);
     
     \draw[->] (s4) --node[above] {\scriptsize $a$} (acc);
     \draw[dashed, ->] (s4) --node[below] {\scriptsize $b$} (rej);
     
     \draw[dashed, ->] (s5) -- (acc);
     \draw[->] (s5) --node[below] {\scriptsize $a$} (rej);

    \end{tikzpicture}}}
\subcaptionbox{Example for shields    \label{fig:shieldconvergence}}{
    \scalebox{1.0}{
    \begin{tikzpicture}
     \node[circle,draw, initial, initial text=, initial where=above] (s1) {};
     \node [cloud, draw,cloud puffs=10,cloud puff arc=120, aspect=1.7, inner ysep=0.3em,left=0.3cm of s1, fill=black!10] (cx) {\scriptsize $A$};
     \node [cloud, draw,cloud puffs=10,cloud puff arc=120, aspect=1.7, inner ysep=0.3em,right=0.3cm of s1, fill=black!10] (cy) {\scriptsize $B$};
     \node [cloud, draw,cloud puffs=10,cloud puff arc=120, aspect=1.7, inner ysep=0.3em,below=0.1cm of cy, fill=black!10] (cz) {\scriptsize $C$};

     \node[draw, circle, right=0.4cm of cy,accepting] (acc) {};
     \node[draw, circle, right=0.4cm of cz] (rej) {};

     \draw[->] (s1) -- node[above] {\scriptsize $a$} (cx);
     \draw[->] (s1) -- node[above] {\scriptsize $b$} (cy);
     \draw[->] (s1) -- node[below] {\scriptsize $c$} (cz);

     \draw[->] (cy) -- (acc);
     \draw[->] (cz) -- (acc);
     \draw[->] (cz) -- (rej);
     
    \end{tikzpicture}}}
\caption{Figures for illustrating benefits of knowledge interfaces in terms of convergence rate, see \Cref{sec:dataefficiency}.} 
\end{figure}
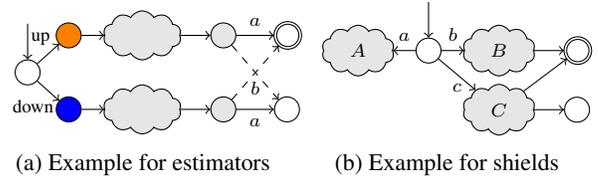

\input{figs/data/evade/noshield_reinforce_belief_supp}
\input{figs/data/evade/shield_reinforce_belief_supp}
\input{figs/data/evade/noshield_reinforce_obs_valuation}
\input{figs/data/evade/shield_reinforce_obs_valuation}

\input{figs/data/avoid/noshield_reinforce_belief_supp}
\input{figs/data/avoid/shield_reinforce_belief_supp}
\input{figs/data/avoid/noshield_reinforce_obs_valuation}
\input{figs/data/avoid/shield_reinforce_obs_valuation}

\input{figs/data/rocks2/noshield_reinforce_belief_supp}
\input{figs/data/rocks2/shield_reinforce_belief_supp}
\input{figs/data/rocks2/noshield_reinforce_obs_valuation}
\input{figs/data/rocks2/shield_reinforce_obs_valuation}

\input{figs/data/intercept/noshield_reinforce_belief_supp}
\input{figs/data/intercept/shield_reinforce_belief_supp}
\input{figs/data/intercept/noshield_reinforce_obs_valuation}
\input{figs/data/intercept/shield_reinforce_obs_valuation}

\input{figs/data/obstacle/noshield_reinforce_belief_supp}
\input{figs/data/obstacle/shield_reinforce_belief_supp}
\input{figs/data/obstacle/noshield_reinforce_obs_valuation}
\input{figs/data/obstacle/shield_reinforce_obs_valuation}

\input{figs/data/refuel/noshield_reinforce_belief_supp}
\input{figs/data/refuel/shield_reinforce_belief_supp}
\input{figs/data/refuel/noshield_reinforce_obs_valuation}
\input{figs/data/refuel/shield_reinforce_obs_valuation}

\begin{figure*}[t]
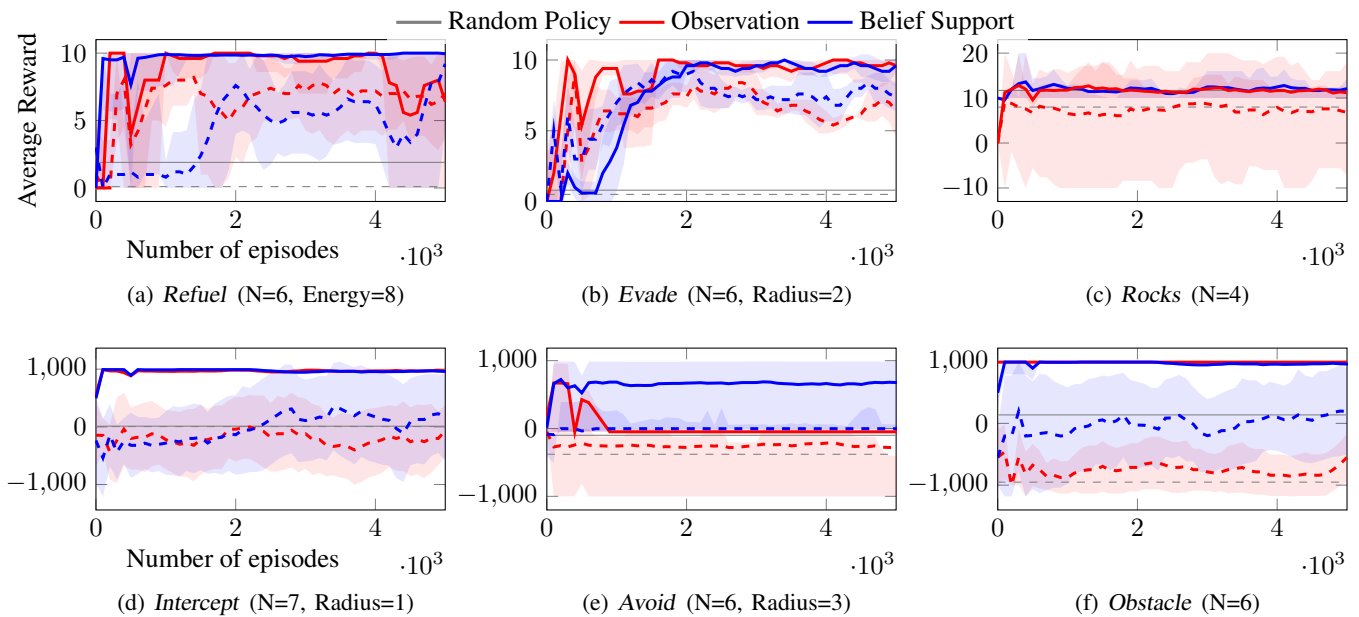

\renewcommand\thesubfigure{(\alph{subfigure})}
\begin{tikzpicture}
\begin{groupplot}[group style={group size=3 by 2,vertical sep=1.95cm,horizontal sep=1.35cm}]
\input{figs/rl_graph_refuel}
\input{figs/rl_graph_evade}
\input{figs/rl_graph_rocks2}
\input{figs/rl_graph_intercept}
\input{figs/rl_graph_avoid}
\input{figs/rl_graph_obstacle}
\end{groupplot}
\node[text width=6cm,align=center,anchor=north] at ([yshift=-7.5mm]group c1r1.south) {\subcaption{\domain{Refuel} (N=6, Energy=8) \label{fig:refuel}}};
\node[text width=6cm,align=center,anchor=north] at ([yshift=-7.5mm]group c2r1.south) {\subcaption{\domain{Evade} (N=6, Radius=2) \label{fig:evade}}};
\node[text width=6cm,align=center,anchor=north] at ([yshift=-7.5mm]group c3r1.south) {\subcaption{\domain{Rocks} (N=4) \label{fig:rocks}}};
\node[text width=6cm,align=center,anchor=north] at ([yshift=-7.5mm]group c1r2.south){\subcaption{\domain{Intercept} (N=7, Radius=1) \label{fig:intercept}}};
\node[text width=6cm,align=center,anchor=north] at ([yshift=-7.5mm]group c2r2.south){\subcaption{\domain{Avoid} (N=6, Radius=3) \label{fig:avoid}}};
\node[text width=6cm,align=center,anchor=north] at ([yshift=-7.5mm]group c3r2.south){\subcaption{\domain{Obstacle} (N=6) \label{fig:obstacle}}};
\end{tikzpicture}
\caption{REINFORCE {with} (solid) and {without} (dashed) a shield. The red lines show when the RL agent is trained using only observations from the environment, and the blue lines indicate when the RL agent is trained using the state estimator. The gray lines are the rewards, averaged over multiple evaluations, obtained via a policy that randomly selects from available actions.}
\label{fig:RL_graphs}
\end{figure*}

\subsection{RL Convergence Rate}
\label{sec:dataefficiency}
Beyond providing safety guarantees, learning in partially observable settings remains a challenge, especially when rewards are sparse. The availability of a partial model provides potential to accelerate the learning process.

\paragraph{Using state estimates.}
A state estimator enriches the observation with a signal that compresses the history. Consider the POMDP sketch in Fig.~\ref{fig:estimatorconvergence}, abstracting a setting where the agent early on makes an observation (\textit{orange}, top) or (\textit{blue}, bottom), must learn to remember this observation until the end, where it has to take either action $a$ (solid) when it saw \textit{orange} before, or action $b$ (dashed) when it saw \textit{blue} before. State estimation provides a belief support that clarifies  whether we are in the bottom or top part of the model, and thus trivializes the learning.

\paragraph{Using shields.}
A shield may provide incentives to (not) explore parts of the state space. Consider an environment as sketched out in Fig.~\ref{fig:shieldconvergence}. 
We have partitioned the state space into three disjoint parts.
In region $A$, there are no bad states (with a high negative reward) but neither are there any reach states, thus, region $A$ is a dead-end.
In region $B$, all states will eventually reach, and in region $C$, there is a (small) probability that we eventually reach an avoid state. 
An agent has to learn that it should always enter region $B$ from the initial state. 
However, if it (uniformly) randomly chooses actions (as an RL agent may do initially) it will only explore region $B$ in one third of the episodes.
If the high negative reward is not encountered early, it will take quite some time to skew the distribution towards entering region $B$. Even worse, in cases where the back-propagation of the sparse reward is slow, region $A$ will remain to appear attractive and region $C$ may appear more attractive whenever back-propagation is faster. 
The latter happens if the paths towards positive reward in region $C$ are shorter than in region $B$.

\subsection{Bootstrapping: Learning From the Shield}
Finally, it is interesting to consider the possibility of disabling the additional interfaces after an initial training phase.
For example, this allows us to bootstrap an agent with the shield and then relax the restrictions it imposes.
Such a setting is relevant whenever the shield is overly conservative -- e.g., entering some avoid-states is unfortunate but not safety-critical.  
It may also simplify the (formal) analysis of the RL agent, e.g., via neural network verification~\cite{katz2017reluplex,carr2021task}, as there is no further need to integrate the shield or state estimator in these analyses.
We investigate two ways to disable these interfaces and to evaluate agent performance after this intervention: either a \emph{smooth transition} or \emph{sudden deactivation}. 

When switching off shields \emph{suddenly}, the agent will be overly reliant on the effect of the shield. While it remembers some good decisions, it must learn to avoid some unsafe actions. 
We want to encourage the agent to learn to not rely on the shield.
To support this idea, we propose a \emph{smooth} transition: When switching off the shield, we do so gradually, applying the shield with some probability $p$, which will allow negative rewards whenever an action not allowed by the shield is taken. We decay the probability that the shield is applied over time to gently fade out its the effect.
\footnote{
When switching off state estimators, the learned agent is no longer executable as it lacks necessary information. 
Solutions, e.g., defaulting to a fixed observation, are not part of this work.}
\section{Experiments}

We applied shielded RL in six challenging domains with partial observability. We compare multiple state-of-the-art deep RL agents.
The experiments focus on the \emph{safety}, \emph{convergence rate}, and the \emph{ability to learn} from additional interfaces as outlined in the three subsections of Section~\ref{sec:benefits}. 
We include the source code, the full set of results, and plots for all learning methods and domains in the technical appendix.

\paragraph{Setup.}
We use \emph{Storm}~\citep{DBLP:journals/sttt/HenselJKQV22} as framework to interface the model, the shield, and the state estimator.
All shields are computed within few minutes, details are given in the Appendix~\ref{sec:Domains}.
We developed bindings to Tensorflow's \emph{TF-Agents} package~\citep{TFAgents} and use its masking function to implement the precomputed shield. 
All experiments were performed on 8x3.2GHz Intel Xeon Platinum 8000 series processor with 32GB of RAM.

We employ a set of grid-based scenarios from \citep{DBLP:conf/cav/JungesJS20}. \domain{Refuel} and \domain{Obstacle} involve guiding a noisy agent to a goal location while avoiding hazardous situations. 
\domain{Avoid} and \domain{Evade} aim to guide an agent to a goal while avoiding collisions with one or more moving robots.
\domain{Intercept} attempts to prevent an adversary escaping by catching it in time. Finally, \domain{Rocks} is a variant of RockSample~\cite{smith2004heuristic} where the agent collects valuable rocks and delivers them. Detailed descriptions of the environments are given in Appendix~\ref{sec:Domains}.

%
%

We use five deep RL methods: DQN~\citep{DBLP:journals/nature/MnihKSRVBGRFOPB15}, DDQN~\citep{DBLP:conf/aaai/HasseltGS16}, PPO~\citep{DBLP:journals/corr/SchulmanWDRK17}, discrete SAC \citep{DBLP:journals/corr/abs-1910-07207} and REINFORCE~\citep{DBLP:journals/ml/Williams92}. 
Unless otherwise specified, we limited episodes to a maximum of 100 steps and calculated the average reward across 10 evaluation episodes.
Due to the sparse reward nature of the domains and for the sake of readability, we performed smoothing for all figures across a five-interval window.
In episodal RL algorithms, such as REINFORCE, we trained on 5000 episodes with an evaluation interval every 100 episodes, and in the step-based RL algorithms, such as DQN, DDQN, PPO and discrete SAC, we trained on $10^5$~steps with an evaluation interval every $1000$~steps. Additionally, in the discrete SAC, we use long short-term memory (LSTM) as comparison to recent LSTM-based deep RL methods on POMDPs~\citep{wierstra2007solving,hausknecht2015deep}. Details on the hyperparameters and the selection method are given in  Appendix~\ref{sec:Hyperparam}.

\subsection{Main Results}
We make some key observations for REINFORCE. Results in Sec.~\ref{sec:DetailedResults} and Appendix~\ref{sec:MoreResults} clarify that, in general, these observations hold for other RL algorithms.

The \emph{no shield} and \emph{shield} rows in Tab.~\ref{tab:violate_numbers} demonstrate that in line with the formal guarantees, \textbf{(only) the shielded agents never violate the safety specification}.

\begin{table}[t]
\centering
	\begin{tabular}{lrr}
		\hline
		\multirow{2}{4cm}{Learning Setting} & \multicolumn{2}{c}{No. violations}\\
		&  \multicolumn{1}{r}{\emph{During}} & \multicolumn{1}{r}{\emph{After}} \\
		\hline
		\emph{No shield} & 3153 & 1023 \\
		\emph{Shield} & 0  & 0 \\
		\emph{Sudden switch-off} & 1867 & 502 \\
		\emph{Smooth switch-off} & 27 & 5\\
		\hline
	\end{tabular}
	\caption{The number of violations per episode for REINFORCE learning agent \textit{during} and \textit{after} learning across 5000 episodes, averaged across the six domains. An RL agent can have, at most, one violation per episode.}
	\label{tab:violate_numbers}
\end{table}

In Fig.~\ref{fig:RL_graphs}, we demonstrate the performance of REINFORCE. In these and subsequent plots, the dashed lines indicate RL agents learning without the shield, while solid lines indicate that the agent is shielded. \textbf{Shielding accelerates convergence.} In Fig.~\ref{fig:avoid} \&~\ref{fig:obstacle} we observe that the addition of \textbf{the state estimator} (blue) \textbf{improves the convergence rate} over simply having the agent attempt to learn from observations (red).  
As the presented domains are partially observable with sparse reward, they are challenging settings for RL. Consequently, we see that REINFORCE does not always converge. We discuss details in Sec.~\ref{sec:DetailedResults}.

\begin{figure}[t]
	\centering
	\input{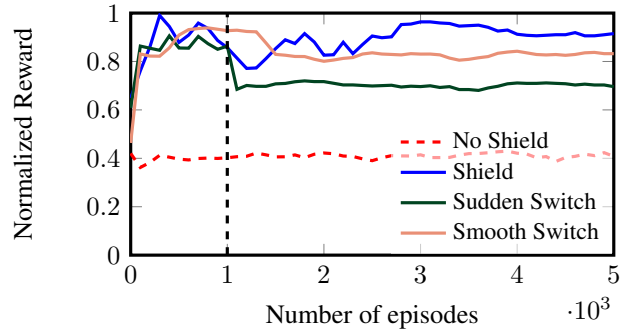}
	\caption[Shield switching]{Normalized reward across all domains for RL agent (with the state estimator)$^{\ref{ft:state_est}}$ that learns for the first 1000 episodes with the shield active. After 1000 episodes the shield is either switched off completely (green) or is slowly turned off with increasing probability (yellow).}
	\label{fig:ShieldSwitchNorm}
\end{figure}

In Fig.~\ref{fig:ShieldSwitchNorm}, we show how an RL agent performs when it initially learns using a shield and then that shield is either completely deactivated after 1000 episodes (green) or is switched-off with a smooth transition (orange). For the latter, we apply the shield with probability $p$ and reduce $p$ from $1$ to $0$ by learning rate $\alpha$.
The shielded RL agent generates higher-quality episodes and subsequently, \textbf{when the shield is removed, the agent still maintains higher quality episodes} since it has previously experienced the sparse positive reward. The effect is even more pronounced as the shield is smoothly removed, where the performance mirrors the shielded condition. We also refer to Tab.~\ref{tab:violate_numbers} for aggregates on safety violations when switching off shields. The \textbf{smooth deactivation cannot prevent safety violations completely, but shows a considerable improvement over the unshielded version}.

\subsection{Detailed Results}
\label{sec:DetailedResults}
In the sequel, we highlight important elements of the challenges presented in sparse domains, the shield's improved performance, and how the belief support and its representation impact learning.

\begin{figure}[t]
    \centering
    \input{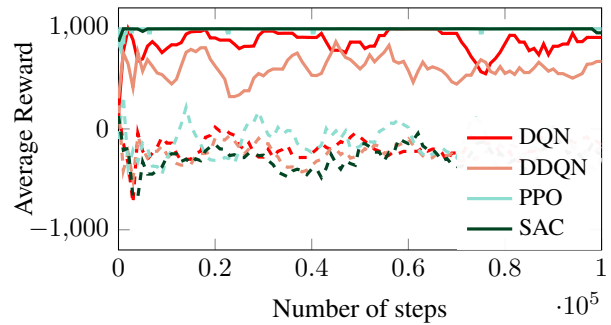}
    \caption[Different learning methods for \domain{Intercept}]{RL agents$^{\ref{ft:state_est}}$ employing different learning methods for \domain{Intercept}.}
    \label{fig:rl_methods}
\end{figure}

\begin{figure}[t]
	\centering
	\input{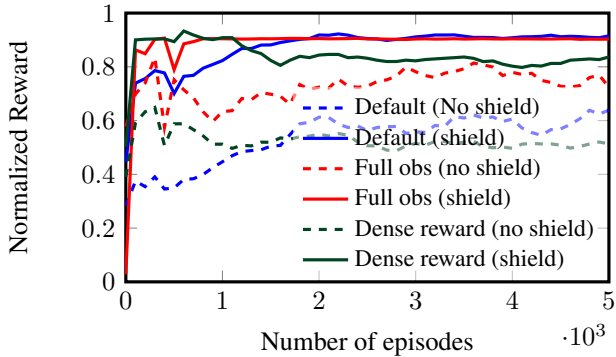}
	\caption[Empirical study]{Empirical study comparing the different inputs for performing REINFORCE on a reward normalized across all domains. Dashed lines imply the RL agent$^{\ref{ft:state_est}}$ is not shielded.}
	\label{fig:ablation_norm}
\end{figure}

\paragraph{Domains are sparse and thus challenging.}
As discussed, Fig.~\ref{fig:RL_graphs} \& Fig.~\ref{fig:ablation_norm} indicate that in the sparse-reward domains and under partial observation, without using additional knowledge from the given partial model, the deep RL algorithms struggle.
In particular, reaching target states with a random policy is very unlikely, e.g., in \domain{Evade} (Fig.~\ref{fig:evade}), a random policy without a shield reaches the target approximately $1\%$ of the time. 
Likewise, when the agent attempts to learn a policy for \domain{Avoid}, it converges to a locally optimal but globally sub-optimal policy, with an average reward of $-100$ (global optimum of $+991$).
This policy that remains in the left corner stays outside of the adversary's reachable space, but will not move towards the goal at all.
Similarly, the unshielded random policy often reaches a highly negative reward: e.g., $95\%$ of the time in \domain{Obstacle} (Fig.~\ref{fig:obstacle}).
In POMDP settings converging to a local optimum is a challenge for many RL agents: In Fig.~\ref{fig:rl_methods}, we illustrate the problematic performance on the \domain{Intercept} domain for a variety of unshielded RL agents.

\paragraph{Ablation study: full observability and reward shaping.} 
In Fig.~\ref{fig:ablation_norm}, we investigate the RL agent's\footnote{\label{ft:state_est}Each RL agent used the belief support (via the state estimator) for the policy input representation.} performance in more detail. In particular, when artificially making the domain fully observable, REINFORCE learns the optimal policy quickly for all domains (even in the unshielded condition), which demonstrates how difficult it is to learn in POMDPs. To overcome the sparse reward we can use reward shaping to guide the learner~\cite{DBLP:conf/aaai/KimLLNK15,DBLP:conf/ijcci/HlynssonW21}. While reward shaping\footnote{Details for specific shaping in Fig.~\ref{fig:ablation_norm} are in Appendix~\ref{sec:reward_shape}.} may help, it leads to nontrivial side-effects, especially for POMDPs. 
We observe that a dense reward function helps the RL agent to converge faster than in the default domain (see first 1000 episodes in Fig.~\ref{fig:ablation_norm}). However, the dense reward may harm the final performance, as exemplified by the performance for the dense reward compared to the default.  
In particular, it seems to encourage risky exploration in unshielded settings.  
The shield provides the best improvement of performance when viewed in isolation (each solid line gives consistently higher returns than its dashed equivalent).

\paragraph{Shields improve convergence rate.}
Shielded agents prevent encountering avoid states in all episodes, and other episodes are thus more frequent. 
Consequently, a shielded RL agent has a higher probability of obtaining a positive reward even if the reward is sparse.
For instance, in \domain{Obstacle}, an unshielded random policy averages approximately 12 steps before crashing. In contrast, the shielded policy, which cannot crash, averages approximately 47 steps before reaching the goal. For RL agents that rely on episodic training, such as REINFORCE, the shield greatly improves the agent's convergence rate, see Fig.~\ref{fig:obstacle}. These efficiencies hold even for the step-based RL agents, such as those presented in Fig.~\ref{fig:rl_methods}. However, the DQN and DDQN struggle to converge to the optimal policy. Such behavior could result from insufficient data to properly process the state estimates from the shield.

\paragraph{State estimators improve convergence (less).}
The challenge of RL agents struggling with high uncertainty, as sketched in the previous paragraphs, 
can also occur with a shield.
Again, in the \domain{Obstacle} domain, 
REINFORCE without the state estimation (red in Fig.~\ref{fig:RL_graphs})
needs to learn both how to map the observation to the possible states, and then also how this would map into a value function, which it does only after spending roughly 2000 episodes. 
In comparison, with access to the belief support (blue in Fig.~\ref{fig:RL_graphs}), the agent quickly learns to estimate the value function. Thus, even shielded, access to a state estimator can help. 
Vice versa, a shield does significantly improve agents, even with a state estimator.



\paragraph{Limitation: Shields alone do not enforce reaching targets quickly.}
As a drawback, shielding does not directly steer the agent towards a positive reward. 
In environments like \domain{Evade}, where the reward is particularly sparse, a random policy with a shield has only an $8\%$ chance of reaching the goal, see Fig.~\ref{fig:evade}.
In particular, it takes many episodes before even collecting any positive reward.
Shielded agents do thus not alleviate the fact that episodes may need to be long.

\begin{figure}[t]
    \centering
    \renewcommand\thesubfigure{(\alph{subfigure})}
    \subcaptionbox{\emph{Evade} at $t=9$\label{fig:badshield_A}}[1.25in]{\includegraphics[width=\screenshotsize,height=\screenshotsize]{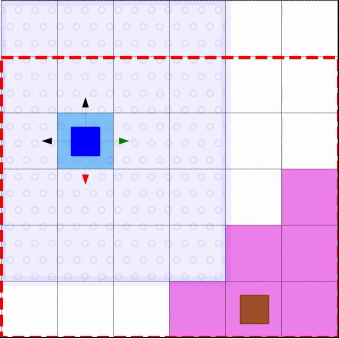}}\qquad
    \subcaptionbox{\emph{Evade} at $t=25$\label{fig:badshield_B}}[1.25in]{\includegraphics[width=\screenshotsize,height=\screenshotsize]{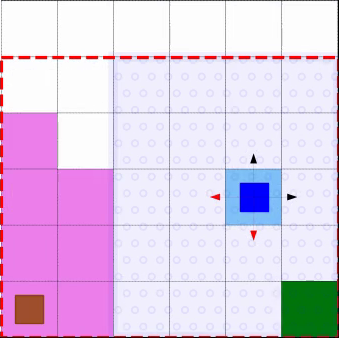}}
    \caption{Incremental states of \emph{Evade} 
    where the agent (dark blue square) has a belief set of states (shaded in pink). The goal (green) is static. At $t=9$, the shield prevents $\{\textrm{south}\}$ and the agent takes $\{\textrm{east}\}$ and at $t=25$, the shield prevents $\lbrace\textrm{south},\textrm{east}\rbrace$ and the agent takes $\{\textrm{scan}\}$.}
    \label{fig:bad_shield}
\end{figure}

\paragraph{Limitation: Shields may have little effect on performance.}
For the domain \domain{Evade} in Fig.~\ref{fig:evade}, the RL agent is only marginally improved by the addition of the shield. 
In this domain, the shield is much less restrictive, often not restricting the agent's choice at all. 
Consider Fig.~\ref{fig:bad_shield}, where the agent can easily either take an action that is just as beneficial as the one that was restricted as in Fig.~\ref{fig:badshield_A}, or reduce the size of the belief support by taking a scan as in Fig.~\ref{fig:badshield_B}. Further, in \emph{Evade}, the shield is restricting the agent from taking actions that result in collisions with a very low probability. When the unshielded agent takes these potentially unsafe actions, it rarely suffers any negative outcome, leading to similar values of average reward. In principle, this may even degrade the performance of the shield. 
A similar problem occurs if the episodes are too short to ensure reaching the target, as detailed in Appendix~\ref{sec:VariableLength}.

\section{Conclusions}
We presented a thorough study and an efficient open-source integration of model-based shielding and data-driven RL towards safe 
learning in partially observable settings. 
The shield ensures that the RL agent never visits dangerous avoid-states or dead-ends.
Additionally, the use of shields helps to accelerate state-of-the-art RL. 
For future work, we will investigate the use of model-based distance measures to target states~\citep{jansen-et-al-concur-2020} or contingency plans \citep{DBLP:journals/jair/PryorC96,DBLP:conf/aips/BertoliCP06} as an additional interface to the agent.

\bibliography{literature}

\clearpage
\appendix
\clearpage

\input{figs/data/evade/noshield_reinforce_belief_supp}
\input{figs/data/evade/shield_reinforce_belief_supp}
\input{figs/data/evade/noshield_reinforce_obs_valuation}
\input{figs/data/evade/shield_reinforce_obs_valuation}

\input{figs/data/avoid/noshield_reinforce_belief_supp}
\input{figs/data/avoid/shield_reinforce_belief_supp}
\input{figs/data/avoid/noshield_reinforce_obs_valuation}
\input{figs/data/avoid/shield_reinforce_obs_valuation}

\input{figs/data/rocks2/noshield_reinforce_belief_supp}
\input{figs/data/rocks2/shield_reinforce_belief_supp}
\input{figs/data/rocks2/noshield_reinforce_obs_valuation}
\input{figs/data/rocks2/shield_reinforce_obs_valuation}

\input{figs/data/intercept/noshield_reinforce_belief_supp}
\input{figs/data/intercept/shield_reinforce_belief_supp}
\input{figs/data/intercept/noshield_reinforce_obs_valuation}
\input{figs/data/intercept/shield_reinforce_obs_valuation}

\input{figs/data/obstacle/noshield_reinforce_belief_supp}
\input{figs/data/obstacle/shield_reinforce_belief_supp}
\input{figs/data/obstacle/noshield_reinforce_obs_valuation}
\input{figs/data/obstacle/shield_reinforce_obs_valuation}

\input{figs/data/refuel/noshield_reinforce_belief_supp}
\input{figs/data/refuel/shield_reinforce_belief_supp}
\input{figs/data/refuel/noshield_reinforce_obs_valuation}
\input{figs/data/refuel/shield_reinforce_obs_valuation}

\input{figs/data/evade/noshield_ddqn_belief_supp}
\input{figs/data/evade/shield_ddqn_belief_supp}
\input{figs/data/evade/noshield_ddqn_obs_valuation}
\input{figs/data/evade/shield_ddqn_obs_valuation}

\input{figs/data/avoid/noshield_ddqn_belief_supp}
\input{figs/data/avoid/shield_ddqn_belief_supp}
\input{figs/data/avoid/noshield_ddqn_obs_valuation}
\input{figs/data/avoid/shield_ddqn_obs_valuation}

\input{figs/data/rocks2/noshield_ddqn_belief_supp}
\input{figs/data/rocks2/shield_ddqn_belief_supp}
\input{figs/data/rocks2/noshield_ddqn_obs_valuation}
\input{figs/data/rocks2/shield_ddqn_obs_valuation}

\input{figs/data/intercept/noshield_ddqn_belief_supp}
\input{figs/data/intercept/shield_ddqn_belief_supp}
\input{figs/data/intercept/noshield_ddqn_obs_valuation}
\input{figs/data/intercept/shield_ddqn_obs_valuation}

\input{figs/data/obstacle/noshield_ddqn_belief_supp}
\input{figs/data/obstacle/shield_ddqn_belief_supp}
\input{figs/data/obstacle/noshield_ddqn_obs_valuation}
\input{figs/data/obstacle/shield_ddqn_obs_valuation}

\input{figs/data/refuel/noshield_ddqn_belief_supp}
\input{figs/data/refuel/shield_ddqn_belief_supp}
\input{figs/data/refuel/noshield_ddqn_obs_valuation}
\input{figs/data/refuel/shield_ddqn_obs_valuation}

\input{figs/data/evade/noshield_dqn_belief_supp}
\input{figs/data/evade/shield_dqn_belief_supp}
\input{figs/data/evade/noshield_dqn_obs_valuation}
\input{figs/data/evade/shield_dqn_obs_valuation}

\input{figs/data/avoid/noshield_dqn_belief_supp}
\input{figs/data/avoid/shield_dqn_belief_supp}
\input{figs/data/avoid/noshield_dqn_obs_valuation}
\input{figs/data/avoid/shield_dqn_obs_valuation}

\input{figs/data/rocks2/noshield_dqn_belief_supp}
\input{figs/data/rocks2/shield_dqn_belief_supp}
\input{figs/data/rocks2/noshield_dqn_obs_valuation}
\input{figs/data/rocks2/shield_dqn_obs_valuation}

\input{figs/data/intercept/noshield_dqn_belief_supp}
\input{figs/data/intercept/shield_dqn_belief_supp}
\input{figs/data/intercept/noshield_dqn_obs_valuation}
\input{figs/data/intercept/shield_dqn_obs_valuation}

\input{figs/data/obstacle/noshield_dqn_belief_supp}
\input{figs/data/obstacle/shield_dqn_belief_supp}
\input{figs/data/obstacle/noshield_dqn_obs_valuation}
\input{figs/data/obstacle/shield_dqn_obs_valuation}

\input{figs/data/refuel/noshield_dqn_belief_supp}
\input{figs/data/refuel/shield_dqn_belief_supp}
\input{figs/data/refuel/noshield_dqn_obs_valuation}
\input{figs/data/refuel/shield_dqn_obs_valuation}

\input{figs/data/evade/noshield_sac_belief_supp}
\input{figs/data/evade/shield_sac_belief_supp}
\input{figs/data/evade/noshield_sac_obs_valuation}
\input{figs/data/evade/shield_sac_obs_valuation}

\input{figs/data/avoid/noshield_sac_belief_supp}
\input{figs/data/avoid/shield_sac_belief_supp}
\input{figs/data/avoid/noshield_sac_obs_valuation}
\input{figs/data/avoid/shield_sac_obs_valuation}

\input{figs/data/rocks2/noshield_sac_belief_supp}
\input{figs/data/rocks2/shield_sac_belief_supp}
\input{figs/data/rocks2/noshield_sac_obs_valuation}
\input{figs/data/rocks2/shield_sac_obs_valuation}

\input{figs/data/intercept/noshield_sac_belief_supp}
\input{figs/data/intercept/shield_sac_belief_supp}
\input{figs/data/intercept/noshield_sac_obs_valuation}
\input{figs/data/intercept/shield_sac_obs_valuation}

\input{figs/data/obstacle/noshield_sac_belief_supp}
\input{figs/data/obstacle/shield_sac_belief_supp}
\input{figs/data/obstacle/noshield_sac_obs_valuation}
\input{figs/data/obstacle/shield_sac_obs_valuation}

\input{figs/data/refuel/noshield_sac_belief_supp}
\input{figs/data/refuel/shield_sac_belief_supp}
\input{figs/data/refuel/noshield_sac_obs_valuation}
\input{figs/data/refuel/shield_sac_obs_valuation}

\input{figs/data/evade/noshield_ppo_belief_supp}
\input{figs/data/evade/shield_ppo_belief_supp}
\input{figs/data/evade/noshield_ppo_obs_valuation}
\input{figs/data/evade/shield_ppo_obs_valuation}

\input{figs/data/avoid/noshield_ppo_belief_supp}
\input{figs/data/avoid/shield_ppo_belief_supp}
\input{figs/data/avoid/noshield_ppo_obs_valuation}
\input{figs/data/avoid/shield_ppo_obs_valuation}

\input{figs/data/rocks2/noshield_ppo_belief_supp}
\input{figs/data/rocks2/shield_ppo_belief_supp}
\input{figs/data/rocks2/noshield_ppo_obs_valuation}
\input{figs/data/rocks2/shield_ppo_obs_valuation}

\input{figs/data/intercept/noshield_ppo_belief_supp}
\input{figs/data/intercept/shield_ppo_belief_supp}
\input{figs/data/intercept/noshield_ppo_obs_valuation}
\input{figs/data/intercept/shield_ppo_obs_valuation}

\input{figs/data/obstacle/noshield_ppo_belief_supp}
\input{figs/data/obstacle/shield_ppo_belief_supp}
\input{figs/data/obstacle/noshield_ppo_obs_valuation}
\input{figs/data/obstacle/shield_ppo_obs_valuation}

\input{figs/data/refuel/noshield_ppo_belief_supp}
\input{figs/data/refuel/shield_ppo_belief_supp}
\input{figs/data/refuel/noshield_ppo_obs_valuation}
\input{figs/data/refuel/shield_ppo_obs_valuation}
\section{Technical Appendix}

In this section, we present the setup for the experiments. In particular we give details for the set of experiment domains, reward models and RL agent hyperparameters.

\subsection{Domain Descriptions}
\label{sec:Domains}
\paragraph{\domain{Rocks}}
\domain{Rocks} is a variant of \emph{RockSample}~\citep{smith2004heuristic}.
The grid contains two rocks which are either valuable or dangerous to collect. 
To find out with certainty, the rock has to be sampled from an adjacent field.
The goal is to collect a valuable rock (+10 reward), bring it to the drop-off zone (+10), and not collect dangerous rocks (-10).
\paragraph{\domain{Refuel}}
\domain{Refuel} concerns a rover that shall travel from one corner to the other (+10 reward), while avoiding an obstacle on the diagonal. Every movement costs energy, and the rover may recharge at dedicated stations to its full battery capacity, but neither action yields a reward or cost.  Collisions and empty battery levels terminate the episode.
The rover receives noisy information about its position and battery level.
\paragraph{\domain{Evade}}
\domain{Evade} is a scenario where an agent needs to reach an escape door (+10 reward) and evade a faster robot. 
The agent has a limited range of vision ($\textrm{Radius}$), but may choose to scan the whole grid instead of moving.
\paragraph{\domain{Avoid}}
\domain{Avoid} is a related scenario where an agent attempts to reach a goal (+1000) in the opposite corner and keep a distance from patrolling robots on fixed routes that move with uncertain speed, yielding partial information about their position. If being caught, the robot receives a reward of (-1000). Furthermore, every step yields -1 reward.

\begin{table}[h!]
\centering
    \scalebox{1.0}{
        \begin{tabular}{lll}
            \hline
            \multirow{4}{*}{\domain{Rocks}}
            & \multicolumn{1}{l}{Episode Length} & \multicolumn{1}{l}{100}\\\cline{2-3}
            & \multicolumn{1}{l}{Grid-size} & \multicolumn{1}{l}{6} \\\cline{2-3}
            & \multicolumn{1}{l}{States $|\states|$} & 331 \\\cline{2-3}
            & \multicolumn{1}{l}{Shield compute time (s)} &  19 \\
            \hline
            \multirow{5}{*}{\domain{Refuel}}
            & \multicolumn{1}{l}{Episode Length} & \multicolumn{1}{l}{100}\\\cline{2-3}
            & \multicolumn{1}{l}{Grid-size} & \multicolumn{1}{l}{6} \\\cline{2-3}
            & \multicolumn{1}{l}{Maximum energy level} & \multicolumn{1}{l}{8} \\\cline{2-3}
            & \multicolumn{1}{l}{States $|\states|$} & 270 \\\cline{2-3}
            & \multicolumn{1}{l}{Shield compute time (s)} & 6 \\
            \hline
            \multirow{5}{*}{\domain{Evade}}
            & \multicolumn{1}{l}{Episode Length} & \multicolumn{1}{l}{350}\\\cline{2-3}
            & \multicolumn{1}{l}{Grid-size} & \multicolumn{1}{l}{6} \\\cline{2-3}
            & \multicolumn{1}{l}{Vision radius} & \multicolumn{1}{l}{2} \\\cline{2-3}
            & \multicolumn{1}{l}{States $|\states|$} & 4232 \\\cline{2-3}
            & \multicolumn{1}{l}{Shield compute time (s)} & 142 \\
            \hline
            \multirow{5}{*}{\domain{Avoid}}
            & \multicolumn{1}{l}{Episode Length} & \multicolumn{1}{l}{100}\\\cline{2-3}
            & \multicolumn{1}{l}{Grid-size} & \multicolumn{1}{l}{100} \\\cline{2-3}
            & \multicolumn{1}{l}{Vision radius} & \multicolumn{1}{l}{3} \\\cline{2-3}
            & \multicolumn{1}{l}{States $|\states|$} & 5976 \\\cline{2-3}
            & \multicolumn{1}{l}{Shield compute time (s)} & 167 \\
            \hline
            \multirow{5}{*}{\domain{Intercept}}
            & \multicolumn{1}{l}{Episode Length} & \multicolumn{1}{l}{100}\\\cline{2-3}
            & \multicolumn{1}{l}{Grid-size} & \multicolumn{1}{l}{7} \\\cline{2-3}
            & \multicolumn{1}{l}{Vision radius} & \multicolumn{1}{l}{1} \\\cline{2-3}
            & \multicolumn{1}{l}{States $|\states|$} & 4705 \\\cline{2-3}
            & \multicolumn{1}{l}{Shield compute time  (s)} & 116 \\
            \hline
            \multirow{5}{*}{\domain{Obstacle}}
            & \multicolumn{1}{l}{Episode Length} & \multicolumn{1}{l}{100}\\\cline{2-3}
            & \multicolumn{1}{l}{Grid-size} & \multicolumn{1}{l}{6} \\\cline{2-3}
            & \multicolumn{1}{l}{States $|\states|$} & 37 \\\cline{2-3}
            & \multicolumn{1}{l}{Shield compute time (s)} & 2 \\
            \hline
        \end{tabular}}
    \caption{Constants and parameters for each environment in experimental setups.}
    \label{tab:Environements}
\end{table}

\paragraph{\domain{Intercept}}
Contrary to \domain{Avoid}, in \domain{Intercept} an agent aims to 
meet (+1000) a robot before that robot leaves the grid via one of two available exits (-1000).
The agent has a view radius and observes a corridor in the center of the grid. Movements are penalized with a reward of -1.
\paragraph{\domain{Obstacle}}
\domain{Obstacle} describes an agent navigating through a maze (movement: -1) of static traps where the agent's initial state and movement distance is uncertain, and it only observes whether the current position is a trap (-1000) or exit (+1000).

\subsection{Normalized Reward}
\label{sec:rewardnorm}
To evaluate RL agent performance across all domains we modify the reward such that the agent will take a value between $0$ and $1$, where $1$ is the known highest possible value that the RL agent can achieve. For each domain, this normalization involves the following computations:
\paragraph{\domain{Rocks}}
We add 10 to each return, then divide by 30.

\paragraph{\domain{Refuel}}
We divide each return by 10.

\paragraph{\domain{Evade}}
We divide each return by 10.

\paragraph{\domain{Avoid}}
We add 1000 to each return, then divide by 2000.

\paragraph{\domain{Intercept}}
We add 1000 to each return, then divide by 2000.

\paragraph{\domain{Obstacle}}
We add 1000 to each return, then divide by 2000.

\subsection{Dense Reward Shaping}
\label{sec:reward_shape}
We examine the effect of reward shaping that provides a guide for the RL agent while learning. While these \emph{shaped-reward} functions are provided to the leaner, we perfrom evaluations on the original sparse domains. For each domain, this dense reward shaping takes the following form:
\paragraph{\domain{Rocks}}
The agent receives a reward as a function of distance to the drop-off zone and proximity to a good rock.

\paragraph{\domain{Refuel}}
The agent receives a reward as a function of distance to the target.

\paragraph{\domain{Evade}}
The agent receives a reward as a function of distance to the target.

\paragraph{\domain{Avoid}}
The agent receives a reward as a function of distance to the target.

\paragraph{\domain{Intercept}}
The agent receives a reward as a function of relative distance between itself and the the robot.

\paragraph{\domain{Obstacle}}
The agent receives a reward as a function of distance to the target.

\newpage

\subsection{Hyperparameter Selection}
\label{sec:Hyperparam}
\paragraph{Network parameters}
To study the effect of a shield on different RL methods and domains, we ensured that the chosen hyperparameters were consistent between each experiment. We employed the default settings from the examples provided in the \emph{tf-agents}~\citep{TFAgents} documentation version 0.9.0 with one exception. For discrete SAC~\citep{DBLP:journals/corr/abs-1910-07207}, we modify the \emph{tf-agents}~\cite{TFAgents} implement to handle discrete actions but also we added an LSTM layer in the actor network, see Table~\ref{tab:hyperparam_SAC}. The hyperparameter values for each learning setting are given in~\cref{tab:hyperparam_dqn,tab:hyperparam_drqn,tab:hyperparam_REINFORCE,tab:hyperparam_PPO,tab:hyperparam_SAC}.
Tuning for all domains and agents is beyond the scope of this work.

In \cref{fig:hyperparameter-networksize,fig:hyperparameter-learnrate,fig:hyperparameterobs-networksize,fig:hyperparameterobs-learnrate} we show the effects of varying hyperparameters for the \domain{Refuel} and \domain{Obstacle}.
In \cref{fig:hyperparameter-networksize,fig:hyperparameterobs-networksize} we examine the effect of varying the nodes per layer in the policy network of the REINFORCE algorithm. Similarly, in \cref{fig:hyperparameter-learnrate,fig:hyperparameterobs-learnrate} we examine the effect of varying the learning rate training parameter.

\begin{figure}
    \centering
    \input{figs/Hyperparamters}
    \caption{Performance of training on different values of the policy network's hyperparameters with and without a shield. Each point represents the policy performance on \domain{Refuel} across 10 evaluations after 2000 episodes of REINFORCE.}
    \label{fig:hyperparameter-networksize}
\end{figure}

\begin{figure}
    \centering
     \input{figs/HyperparametersLearn}
    \caption{Performance of training on learning rates for an RL agent with and without a shield. Each point represents the policy performance on \domain{Refuel} across 10 evaluations after 2000 episodes of REINFORCE.}
    \label{fig:hyperparameter-learnrate}
\end{figure}

\begin{figure}
	\centering
	\input{figs/HyperparamtersObstacle}
	\caption{Performance of training on different values of the policy network's hyperparameters with and without a shield. Each point represents the policy performance on \domain{Obstacle} across 10 evaluations after 2000 episodes of REINFORCE.}
	\label{fig:hyperparameterobs-networksize}
\end{figure}

\begin{figure}
	\centering
	\input{figs/HyperparametersObstacleLearn}
	\caption{Performance of training on learning rates for an RL agent with and without a shield. Each point represents the policy performance on \domain{Obstacle} across 10 evaluations after 2000 episodes of REINFORCE.}
	\label{fig:hyperparameterobs-learnrate}
\end{figure}

\begin{table}
	\begin{tabular}{llr}
		\hline
		\multirow{3}{2.5cm}{Actor Network Parameters}
		& \multicolumn{1}{l}{Hidden layers} & \multicolumn{1}{l}{1}\\\cline{2-3}
		& \multicolumn{1}{l}{Nodes per layer} & \multicolumn{1}{l}{100} \\\cline{2-3}
		& \multicolumn{1}{l}{Activation function} & \multicolumn{1}{l}{ReLu} \\
		\hline
		\multirow{4}{2.5cm}{Value Network Parameters}
		& \multicolumn{1}{l}{Hidden layers} & \multicolumn{1}{l}{1}\\\cline{2-3}
		& \multicolumn{1}{l}{Nodes per layer} & \multicolumn{1}{l}{100} \\\cline{2-3}
		& \multicolumn{1}{l}{Activation function} & \multicolumn{1}{l}{ReLu} \\\cline{2-3}
		& \multicolumn{1}{l}{Value Est. Loss Coeff.} & \multicolumn{1}{l}{0.2} \\
		\hline
		\multirow{4}{2.5cm}{Training Parameters}
		& \multicolumn{1}{l}{Optimizer} & \multicolumn{1}{l}{ADAM}\\\cline{2-3}
		& \multicolumn{1}{l}{Learning rate} & \multicolumn{1}{l}{\(3e-2\)}\\\cline{2-3}
		& \multicolumn{1}{l}{Minibatch size} & \multicolumn{1}{l}{64} \\\cline{2-3}
		& \multicolumn{1}{l}{Discount $\gamma$} & \multicolumn{1}{l}{1} \\\cline{2-3}
		\hline
		\multirow{2}{*}{Other Parameters}
		& \multicolumn{1}{l}{Evaluation Interval} & \multicolumn{1}{l}{100}\\ \cline{2-3}
		& \multicolumn{1}{l}{Evaluation Episodes} & \multicolumn{1}{l}{10}\\ \cline{2-3}
		\hline
\end{tabular}
\caption{Hyperparameters used in deep REINFORCE numerical experiments.}
\label{tab:hyperparam_REINFORCE}
\end{table}

\begin{table}
	\begin{tabular}{llr}
		\hline
		\multirow{3}{2.5cm}{Q-Network Parameters}
		& \multicolumn{1}{l}{Hidden layers} & \multicolumn{1}{l}{1}\\\cline{2-3}
		& \multicolumn{1}{l}{Nodes per layer} & \multicolumn{1}{l}{100} \\\cline{2-3}
		& \multicolumn{1}{l}{Activation function} & \multicolumn{1}{l}{None} \\
		\hline
		\multirow{4}{2.5cm}{Training Parameters}
		& \multicolumn{1}{l}{Optimizer} & \multicolumn{1}{l}{ADAM}\\\cline{2-3}
		& \multicolumn{1}{l}{Learning rate} & \multicolumn{1}{l}{\(3e-2\)}\\\cline{2-3}
		& \multicolumn{1}{l}{Minibatch size} & \multicolumn{1}{l}{64} \\\cline{2-3}
		& \multicolumn{1}{l}{Discount $\gamma$} & \multicolumn{1}{l}{1} \\\cline{2-3}
		\hline
		\multirow{2}{2.5cm}{Other Parameters}
		& \multicolumn{1}{l}{Evaluation Interval} & \multicolumn{1}{l}{1000}\\ \cline{2-3}
		& \multicolumn{1}{l}{Evaluation Episodes} & \multicolumn{1}{l}{10}\\ \cline{2-3}
		\hline
	\end{tabular}
	\caption{Hyperparameters used in deep Q-network (DQN) and double Q learning (DDQN) numerical experiments.}
	\label{tab:hyperparam_dqn}
\end{table}

\begin{table}
	\begin{tabular}{llr}
		\hline
		\multirow{3}{2.5cm}{Actor Network Parameters}
		& \multicolumn{1}{l}{Hidden layers} & \multicolumn{1}{l}{2}\\\cline{2-3}
		& \multicolumn{1}{l}{Nodes per layer} & \multicolumn{1}{l}{(200,100)} \\\cline{2-3}
		& \multicolumn{1}{l}{Activation function} & \multicolumn{1}{l}{tanh} \\
		\hline
		\multirow{3}{2.5cm}{Value Network Parameters}
		& \multicolumn{1}{l}{Hidden layers} & \multicolumn{1}{l}{2}\\\cline{2-3}
		& \multicolumn{1}{l}{Nodes per layer} & \multicolumn{1}{l}{(200,100)} \\\cline{2-3}
		& \multicolumn{1}{l}{Activation function} & \multicolumn{1}{l}{ReLu} \\
		\hline
		\multirow{4}{2.5cm}{Training Parameters}
		& \multicolumn{1}{l}{Optimizer} & \multicolumn{1}{l}{ADAM}\\\cline{2-3}
		& \multicolumn{1}{l}{Learning rate} & \multicolumn{1}{l}{\(3e-2\)}\\\cline{2-3}
		& \multicolumn{1}{l}{Minibatch size} & \multicolumn{1}{l}{64} \\\cline{2-3}
		& \multicolumn{1}{l}{Discount $\gamma$} & \multicolumn{1}{l}{1} \\\cline{2-3}
		\hline
		\multirow{2}{2.5cm}{Other Parameters}
		& \multicolumn{1}{l}{Evaluation Interval} & \multicolumn{1}{l}{1000}\\ \cline{2-3}
		& \multicolumn{1}{l}{Evaluation Episodes} & \multicolumn{1}{l}{10}\\ \cline{2-3}
		\hline
\end{tabular}
\caption{Hyperparameters used in proximal policy optimization (PPO) numerical experiments.}
\label{tab:hyperparam_PPO}
\end{table}

\begin{table}
\centering
    \scalebox{1.0}{
        \begin{tabular}{llr}
            \hline
            \multirow{4}{2.5cm}{Actor Network Parameters}
            & \multicolumn{1}{l}{Hidden layers} & \multicolumn{1}{l}{3}\\\cline{2-3}
            & \multicolumn{1}{l}{Nodes per layer} & \multicolumn{1}{l}{(400,300)} \\\cline{2-3}
            & \multicolumn{1}{l}{LSTM size} & \multicolumn{1}{l}{40} \\\cline{2-3}
            & \multicolumn{1}{l}{Activation function} & \multicolumn{1}{l}{ReLu} \\
            \hline            
            \multirow{4}{2.5cm}{Critic Network Parameters}
            & \multicolumn{1}{l}{Hidden layers} & \multicolumn{1}{l}{2}\\\cline{2-3}
            & \multicolumn{1}{l}{Nodes per layer} & \multicolumn{1}{l}{300} \\\cline{2-3}
            & \multicolumn{1}{l}{LSTM size} & \multicolumn{1}{l}{40} \\\cline{2-3}
            & \multicolumn{1}{l}{Activation function} & \multicolumn{1}{l}{tanh} \\
            \hline
            \multirow{7}{2.5cm}{Training Parameters}
            & \multicolumn{1}{l}{Optimizer} & \multicolumn{1}{l}{ADAM}\\\cline{2-3}
            & \multicolumn{1}{l}{Learning rate} & \multicolumn{1}{l}{\(3e-2\)}\\\cline{2-3}
            & \multicolumn{1}{l}{Minibatch size} & \multicolumn{1}{l}{64} \\\cline{2-3}
            & \multicolumn{1}{l}{Discount $\gamma$} & \multicolumn{1}{l}{1} \\\cline{2-3}
            & \multicolumn{1}{l}{Importance ratio clipping} & \multicolumn{1}{l}{1} \\ \cline{2-3}
            & \multicolumn{1}{l}{Target Update $\tau$} & \multicolumn{1}{l}{0.05}\\ \cline{2-3}
            & \multicolumn{1}{l}{Target Update Period} & \multicolumn{1}{l}{5}\\
            \hline
            \multirow{2}{2.5cm}{Other Parameters}
            & \multicolumn{1}{l}{Evaluation Interval} & \multicolumn{1}{l}{1000}\\ \cline{2-3}
            & \multicolumn{1}{l}{Evaluation Episodes} & \multicolumn{1}{l}{10}\\\hline
        \end{tabular}}
    \caption{Hyperparameters used in discrete soft actor-critic (SAC) numerical experiments.}
    \label{tab:hyperparam_SAC}
\end{table}

\begin{table}
	
	\begin{tabular}{llr}
		\hline
		\multirow{4}{2.5cm}{Q-Network Parameters}
		& \multicolumn{1}{l}{Hidden dense layers} & \multicolumn{1}{l}{2}\\\cline{2-3}
		& \multicolumn{1}{l}{Nodes per layer} & \multicolumn{1}{l}{(50,20)} \\\cline{2-3}
		& \multicolumn{1}{l}{Activation function} & \multicolumn{1}{l}{ReLu} \\\cline{2-3}
		& \multicolumn{1}{l}{LSTM layer size} & \multicolumn{1}{l}{15}\\
		\hline
		\multirow{4}{2.5cm}{Training Parameters}
		& \multicolumn{1}{l}{Optimizer} & \multicolumn{1}{l}{ADAM}\\\cline{2-3}
		& \multicolumn{1}{l}{Learning rate} & \multicolumn{1}{l}{\(3e-2\)}\\\cline{2-3}
		& \multicolumn{1}{l}{Minibatch size} & \multicolumn{1}{l}{64} \\\cline{2-3}
		& \multicolumn{1}{l}{Discount $\gamma$} & \multicolumn{1}{l}{1} \\\cline{2-3}
		\hline
		\multirow{2}{2.5cm}{Other Parameters}
		& \multicolumn{1}{l}{Evaluation Interval} & \multicolumn{1}{l}{1000}\\ \cline{2-3}
		& \multicolumn{1}{l}{Evaluation Episodes} & \multicolumn{1}{l}{10}\\ \cline{2-3}
		\hline
\end{tabular}
\caption{Hyperparameters used in deep recurrent Q-network (DRQN) in memory comparison experiment.}
\label{tab:hyperparam_drqn}
\end{table}



\clearpage
\section{More Results}
\label{sec:MoreResults}

In this section of the appendix we present some additional observations on safe learning via shielding. First we describe some interesting performance outcomes and later we give the results for the complete set of experiments in the main body.

\subsection{Episode Lengths and Degrading Performance}
\label{sec:VariableLength}

\begin{figure}
    \centering
\input{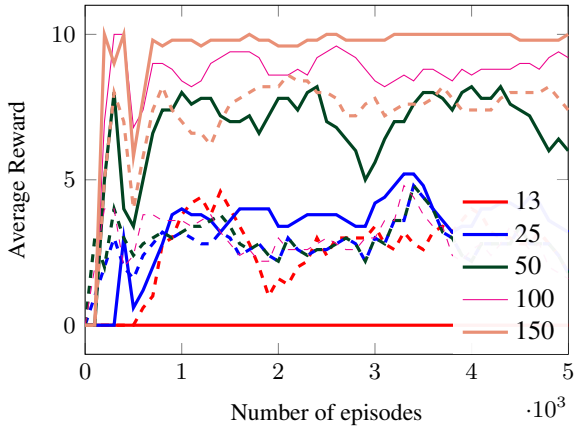}
\caption[Variable episode maximum length for \domain{Refuel}]{Variable episode maximum length for \domain{Refuel}. Solid lines indicate shielded RL agents and dashed lines represent unshielded RL agents.
	\label{fig:learn_var_eps_refuel}}
\end{figure}

Shields can degrade performance.
In Figure~\ref{fig:learn_var_eps_refuel}, we show that in \domain{Refuel}, only when exploring sufficiently long episodes, the agent converges towards an optimal policy. 
In this domain, the agent must rely on the uncertain dynamics to reach the goal without running out of fuel. Just before the possibility of diverting too far from a recharge station, the shield enforces backing up and recharging. It may require several attempts before the agent reaches the goal.
In Figure~\ref{fig:learn_var_eps_refuel} we observe that for (very) short episodes, an unshielded agent may perform better. 
The agent in Figure~\ref{fig:learn_var_eps_refuel} (red dashed) takes the necessary ``risk'' of potentially running out of fuel and using the uncertain dynamics to reach the goal under 13 steps in many (but not all) cases. This violates the safety constraint, but the performance is better than when the (shielded) agent never reaches the goal. This effect fades out with increasing episode length, because the probability that the dynamics turn out favorably increases over time.

\subsection{Fixed Policy For Guiding Learning}
\label{sec:FixedPolicy}
\begin{figure}
 	\centering
 	\input{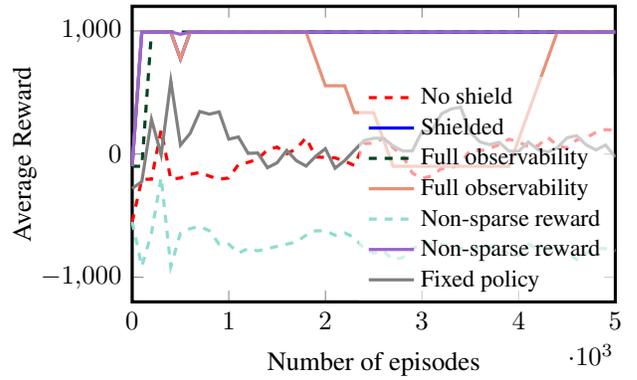}
 	\caption{Ablation study comparing the different inputs for performing REINFORCE on the \domain{Obstacle} domain. Each line used the belief support for the policy input representation.}
 	\label{fig:ablation}
 \end{figure}

We show how giving an RL agent a method of safe exploration can help guide it without necessarily restricting all choices.
We provide an experiment set where the RL agent has a fixed exploration policy, \ie a policy that selects from the shielded set of actions with probability $p$.
The random choice from the safe set of actions means that it explores with more safety than its unshielded counterpart. In Figure~\ref{fig:ablation} Notice the difference in performance in the first 1000 episodes.

\subsection{Input Representation Insights}
\label{sec:InputRep}

\begin{figure}
    \centering
    \hspace{-0.25cm}
    \renewcommand\thesubfigure{(\alph{subfigure})}
    \subcaptionbox{\emph{Obstacle} at $t=2$\label{fig:ambig_A}}[1.1in]{\includegraphics[width=1.0in,height=1.0in]{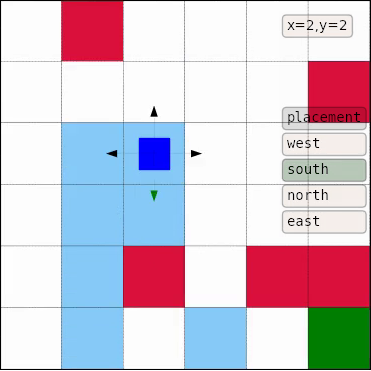}}\hfill
    \subcaptionbox{\emph{Obstacle} at $t=3$\label{fig:ambig_B}}[1.1in]{\includegraphics[width=1.0in,height=1.0in]{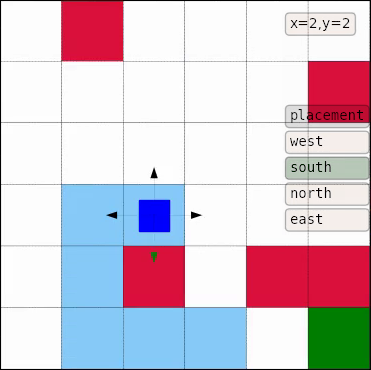}}\hfill
    \subcaptionbox{\emph{Obstacle} at $t=4$\label{fig:ambig_C}}[1.1in]{\includegraphics[width=1.0in,height=1.0in]{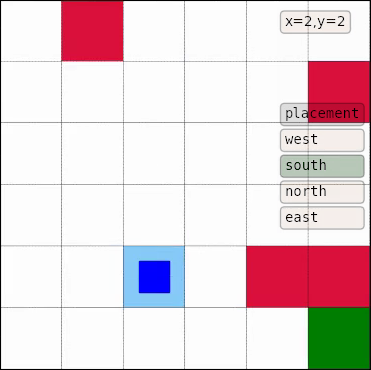}}
    \caption{Incremental states of \domain{Obstacle} environment where the agent (dark blue) handles uncertainty by maintaining a belief set of states (shaded in blue). The goal (green) and obstacles (red) are static. At $t=2$ the agent takes $\textrm{south}$ and again at $t=3$, which results in a collision at $t=4$ }
    \label{fig:ambiguous_states}
\end{figure}

Here we examine different effects of what inputs the RL agent uses to make decisions.
We detail concepts such as partial observability, state estimation, 

\paragraph{Ambiguity in partially observable settings}
One of the challenges of RL in partially observable environments is handling a potentially ambiguous and conflicting set of states. The agent must learn to distinguish states with similar observations. 
This challenge is most evident in the \domain{Obstacle} domain. Consider the agent in Figure~\ref{fig:ambiguous_states}, which could occupy any one of the blue shaded states. At the agent's position at $t=2$ in Figure~\ref{fig:ambig_A}, estimated Q-values (from DQN) are roughly $(733,784,606,687)$ for ($\textrm{west},\textrm{south},\textrm{north},\textrm{east}$) respectively. The unshielded RL agent in this situation is willing to risk possible collision if the agent is in state $x=2$ for the significant advantage gained by taking $\textrm{south}$ for any state in $x=1$. Then, the agent collides with the obstacle at $(x=3,y=4)$, yielding a ${-}1000$ penalty. When the belief support contains just the $x=2$ states, the Q-values are $(499,-456,-417,404)$, which indicates that the DQN algorithm is struggling to account for high uncertainty. 
Shields disable such actions and thus improve further convergence.

\begin{figure}
	\centering
	\input{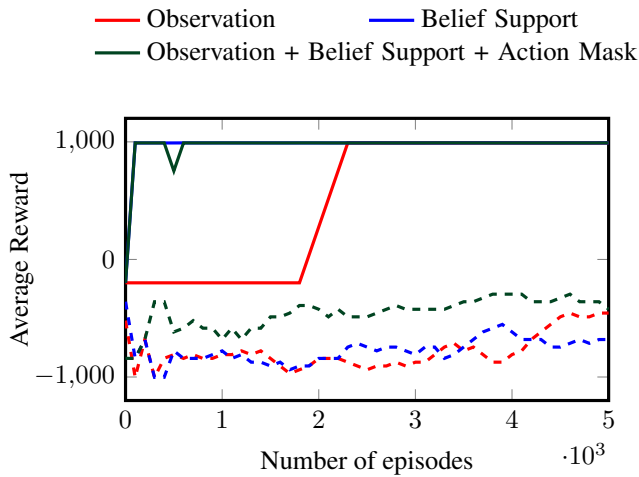}
	\caption{A comparison of three input representations for an RL agent learning on \emph{Obstacle}. The combined representation (green) is an integer vector that contains the information of both the observation vector (red), the belief-support vector (blue) and the action mask at that instant.}
	\label{fig:Stacked_information}
\end{figure}

\paragraph{Input format}
The shield is more than just a state estimate.
In fact, even when we include as much information as possible, in the form of a vector that stacks the observation, the belief-support state estimate and the action mask that a shield would recommend, the shielded RL agent still outperforms its unshielded counterpart. In Figure~\ref{fig:Stacked_information}, a shielded RL agent with a simple observation representation (red) vastly outperforms the unshielded, high-information agent (dashed green).

\begin{figure}
	\input{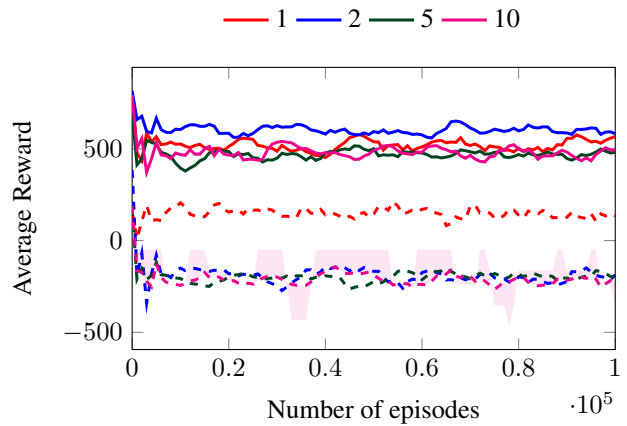}
	\caption{\domain{Intercept} with an LSTM-based SAC agent that interprets sequences of observations through the use of a memory buffer. Each line represents a different instance of how many sequential observations was fed to each agent when learning. See \cite{hausknecht2015deep} for a detailed analysis for the interplay between partially observability and experience replay in RL agents. }
	\label{fig:VariableMemory}
\end{figure}

\paragraph{Experience replay for POMDPs}
For the experience replay, we utilize the uniform sampled replay buffer with a mini-batch size of 64. For DQN, DDQN, PPO and discrete SAC we collect and train in step intervals and for REINFORCE, we collect data as full episode runs.
We also conducted experiments where we gave the RL sequences of observations as an input for training. This experience replay technique is explored in \cite{hausknecht2015deep}, where a RL agent with a DRQN can interpret partial information from multiple observations in sequnce. With that movitation we compared our discrete SAC agent (with its LSTM memory cell) for different input lengths, see Figure~\ref{fig:VariableMemory}. 

\renewcommand\thesubfigure{(\alph{subfigure})}
\clearpage

\begin{figure*}[!h]
	\subsection{Full Observability Data}
	\centering
	\scalebox{1.0}{
\input{figs/full_obs_DQN}}
\caption{Learning using the DQN algorithm in all domains with full observability. }
	\scalebox{1.0}{
\input{figs/full_obs_REINFORCE}}
\caption{Learning using the REINFORCE algorithm in all domains with full observability. }
\end{figure*}

\clearpage
\begin{figure*}[h!]
\subsection{Dense Reward Data}
	\scalebox{1.0}{
		\input{figs/NonSparse_Full_DQN}}
	\caption{Learning using the DQN algorithm in all domains with full observability and a dense reward function to guide learning.}
	\scalebox{1.0}{\input{figs/NonSparse_Full_REINFORCE}}
	\caption{Learning using the REINFORCE algorithm in all domains with full observability and a dense reward function to guide learning.}
\end{figure*}

\begin{figure*}[h!]
		\scalebox{1.0}{\input{figs/NonSparse_Partial_DQN}}
	\caption{Learning using the REINFORCE algorithm in all domains with partial observability and a dense reward function to guide learning.}
		\scalebox{1.0}{\input{figs/NonSparse_Partial_REINFORCE}}
	\caption{Learning using the REINFORCE algorithm in all domains with partial observability and a dense reward function to guide learning.}
\end{figure*}

\clearpage

\begin{figure*}[h!]
	\subsection{Switch Shield Data}
	\input{figs/SWITCH_DQN}
	\caption{Full set of shield switching examples using the DQN learning algorithm.}
	\input{figs/SWITCH_REINFORCE}
	\caption{Full set of shield switching examples using the REINFORCE learning algorithm.}
\end{figure*}

\clearpage
\begin{figure*}[h!]
\subsection{Learning Methods Data}
In \cref{fig:DQN_graphs,fig:DDQN_graphs,fig:PPO_graphs,fig:SAC_graphs}, we show the full set of experiments similar to Figure~3 for REINFORCE.
\vspace{1cm}

\renewcommand\thesubfigure{(\alph{subfigure})}
\begin{tikzpicture}
\begin{groupplot}[group style={group size=3 by 2,vertical sep=1.5cm,horizontal sep=1.35cm}]
\input{figs/AppendixFigs/DQN/rl_graph_refuel}
\input{figs/AppendixFigs/DQN/rl_graph_evade}
\input{figs/AppendixFigs/DQN/rl_graph_rocks2}
\input{figs/AppendixFigs/DQN/rl_graph_intercept}
\input{figs/AppendixFigs/DQN/rl_graph_avoid}
\input{figs/AppendixFigs/DQN/rl_graph_obstacle}
\end{groupplot}
\node[text width=6cm,align=center,anchor=north] at ([yshift=-5.5mm]group c1r1.south) {\subcaption{\domain{Refuel} (N=6, Energy=8)}};
\node[text width=6cm,align=center,anchor=north] at ([yshift=-5.5mm]group c2r1.south) {\subcaption{\domain{Evade} (N=6, Radius=2) }};
\node[text width=6cm,align=center,anchor=north] at ([yshift=-5.5mm]group c3r1.south) {\subcaption{\domain{Rocks} (N=4) }};
\node[text width=6cm,align=center,anchor=north] at ([yshift=-5.5mm]group c1r2.south){\subcaption{\domain{Intercept} (N=7, Radius=1) }};
\node[text width=6cm,align=center,anchor=north] at ([yshift=-5.5mm]group c2r2.south){\subcaption{\domain{Avoid} (N=6, Radius=3) }};
\node[text width=6cm,align=center,anchor=north] at ([yshift=-5.5mm]group c3r2.south){\subcaption{\domain{Obstacle} (N=6) }};
\end{tikzpicture}
\caption{DQN performed with (solid) and without (dashed) a shield restricting unsafe actions. The red lines show when the RL agent is trained using only the observations and the blue lines indicate when the RL agent is trained using some state estimation in the form of belief support. The black lines are the average reward obtained by applying a random policy.}
\label{fig:DQN_graphs}
\end{figure*}

\begin{figure*}
\renewcommand\thesubfigure{(\alph{subfigure})}
\begin{tikzpicture}
\begin{groupplot}[group style={group size=3 by 2,vertical sep=1.5cm,horizontal sep=1.35cm}]
\input{figs/AppendixFigs/DDQN/rl_graph_refuel}
\input{figs/AppendixFigs/DDQN/rl_graph_evade}
\input{figs/AppendixFigs/DDQN/rl_graph_rocks2}
\input{figs/AppendixFigs/DDQN/rl_graph_intercept}
\input{figs/AppendixFigs/DDQN/rl_graph_avoid}
\input{figs/AppendixFigs/DDQN/rl_graph_obstacle}
\end{groupplot}
\node[text width=6cm,align=center,anchor=north] at ([yshift=-5.5mm]group c1r1.south) {\subcaption{\domain{Refuel} (N=6, Energy=8) }};
\node[text width=6cm,align=center,anchor=north] at ([yshift=-5.5mm]group c2r1.south) {\subcaption{\domain{Evade} (N=6, Radius=2) }};
\node[text width=6cm,align=center,anchor=north] at ([yshift=-5.5mm]group c3r1.south) {\subcaption{\domain{Rocks} (N=4) }};
\node[text width=6cm,align=center,anchor=north] at ([yshift=-5.5mm]group c1r2.south){\subcaption{\domain{Intercept} (N=7, Radius=1) }};
\node[text width=6cm,align=center,anchor=north] at ([yshift=-5.5mm]group c2r2.south){\subcaption{\domain{Avoid} (N=6, Radius=3) }};
\node[text width=6cm,align=center,anchor=north] at ([yshift=-5.5mm]group c3r2.south){\subcaption{\domain{Obstacle} (N=6) }};
\end{tikzpicture}
\caption{DDQN performed with (solid) and without (dashed) a shield restricting unsafe actions. The red lines show when the RL agent is trained using only the observations and the blue lines indicate when the RL agent is trained using some state estimation in the form of belief support. The black lines are the average reward obtained by applying a random policy.}
\label{fig:DDQN_graphs}
\end{figure*}

\begin{figure*}
\renewcommand\thesubfigure{(\alph{subfigure})}
\begin{tikzpicture}
\begin{groupplot}[group style={group size=3 by 2,vertical sep=1.5cm,horizontal sep=1.35cm}]
\input{figs/AppendixFigs/PPO/rl_graph_refuel}
\input{figs/AppendixFigs/PPO/rl_graph_evade}
\input{figs/AppendixFigs/PPO/rl_graph_rocks2}
\input{figs/AppendixFigs/PPO/rl_graph_intercept}
\input{figs/AppendixFigs/PPO/rl_graph_avoid}
\input{figs/AppendixFigs/PPO/rl_graph_obstacle}
\end{groupplot}
\node[text width=6cm,align=center,anchor=north] at ([yshift=-5.5mm]group c1r1.south) {\subcaption{\domain{Refuel} (N=6, Energy=8) }};
\node[text width=6cm,align=center,anchor=north] at ([yshift=-5.5mm]group c2r1.south) {\subcaption{\domain{Evade} (N=6, Radius=2) }};
\node[text width=6cm,align=center,anchor=north] at ([yshift=-5.5mm]group c3r1.south) {\subcaption{\domain{Rocks} (N=4)}};
\node[text width=6cm,align=center,anchor=north] at ([yshift=-5.5mm]group c1r2.south){\subcaption{\domain{Intercept} (N=7, Radius=1) }};
\node[text width=6cm,align=center,anchor=north] at ([yshift=-5.5mm]group c2r2.south){\subcaption{\domain{Avoid} (N=6, Radius=3) }};
\node[text width=6cm,align=center,anchor=north] at ([yshift=-5.5mm]group c3r2.south){\subcaption{\domain{Obstacle} (N=6) }};
\end{tikzpicture}
\caption{PPO performed with (solid) and without (dashed) a shield restricting unsafe actions. The red lines show when the RL agent is trained using only the observations and the blue lines indicate when the RL agent is trained using some state estimation in the form of belief support. The black lines are the average reward obtained by applying a random policy.}
\label{fig:PPO_graphs}
\end{figure*}

\begin{figure*}
\renewcommand\thesubfigure{(\alph{subfigure})}
\begin{tikzpicture}
\begin{groupplot}[group style={group size=3 by 2,vertical sep=1.5cm,horizontal sep=1.35cm}]
\input{figs/AppendixFigs/SAC/rl_graph_refuel}
\input{figs/AppendixFigs/SAC/rl_graph_evade}
\input{figs/AppendixFigs/SAC/rl_graph_rocks2}
\input{figs/AppendixFigs/SAC/rl_graph_intercept}
\input{figs/AppendixFigs/SAC/rl_graph_avoid}
\input{figs/AppendixFigs/SAC/rl_graph_obstacle}
\end{groupplot}
\node[text width=6cm,align=center,anchor=north] at ([yshift=-5.5mm]group c1r1.south) {\subcaption{\domain{Refuel} (N=6, Energy=8)}};
\node[text width=6cm,align=center,anchor=north] at ([yshift=-5.5mm]group c2r1.south) {\subcaption{\domain{Evade} (N=6, Radius=2) }};
\node[text width=6cm,align=center,anchor=north] at ([yshift=-5.5mm]group c3r1.south) {\subcaption{\domain{Rocks} (N=4) }};
\node[text width=6cm,align=center,anchor=north] at ([yshift=-5.5mm]group c1r2.south){\subcaption{\domain{Intercept} (N=7, Radius=1) }};
\node[text width=6cm,align=center,anchor=north] at ([yshift=-5.5mm]group c2r2.south){\subcaption{\domain{Avoid} (N=6, Radius=3) }};
\node[text width=6cm,align=center,anchor=north] at ([yshift=-5.5mm]group c3r2.south){\subcaption{\domain{Obstacle} (N=6)}};
\end{tikzpicture}
\caption{Discrete soft-actor critic (SAC) with an LSTM architecture performed with (solid) and without (dashed) a shield restricting unsafe actions. The red lines show when the RL agent is trained using only the observations and the blue lines indicate when the RL agent is trained using some state estimation in the form of belief support.}
\label{fig:SAC_graphs}
\end{figure*}

\end{document}